%% file: Paper_revMF.tex
\documentclass{article}

\usepackage{microtype}
\usepackage{graphicx}
\usepackage{subcaption}
\usepackage{booktabs} % for professional tables
\usepackage{booktabs,tabularx}
\usepackage{booktabs}
\usepackage{longtable}

\usepackage{hyperref}
\usepackage{booktabs}

\usepackage[accepted]{icml2026}

\makeatletter
\renewcommand{\Notice@String}{}
\makeatother

\usepackage{amsmath}
\usepackage{amssymb}
\usepackage{mathtools}
\usepackage{amsthm}

\usepackage[capitalize,noabbrev]{cleveref}

\theoremstyle{plain}

\theoremstyle{definition}

\theoremstyle{remark}

\usepackage[textsize=tiny]{todonotes}

\icmltitlerunning{InterBand-EMG: A Cross-Band Foundation Model for Generalizable Hand-Pose Estimation}

\begin{document}

\twocolumn[
  \icmltitle{InterBand-EMG: A Cross-Band Foundation Model for Generalizable Hand-Pose Estimation}



  \icmlsetsymbol{equal}{*}\icmlsetsymbol{lead}{\textdagger}

  \begin{icmlauthorlist}
    \icmlauthor{Zongheng Guo}{equal,polimi,comp}
    \icmlauthor{Tao Chen}{equal,lead,comp}
    \icmlauthor{Dalong Du}{equal,comp,tsu}
    \icmlauthor{Yijian Yang}{cas,comp}
    \icmlauthor{Heng Liu}{comp}
    \icmlauthor{Lun SHU}{comp}
    \icmlauthor{Chen Luo}{zju,comp}
    \icmlauthor{Junhao Tu}{comp}
    \icmlauthor{Lei Shu}{bit,comp}
    \icmlauthor{Mingzhou Wu}{comp}
    \icmlauthor{Yifan Liu}{comp}
    \icmlauthor{Yu Zhang}{comp}
    \icmlauthor{Zhenqiang Liu}{comp}
    \icmlauthor{Shilei Wen}{comp}
    \icmlauthor{Jie Zhou}{tsu}
    \icmlauthor{Manuela Ferrario}{polimi}

  \end{icmlauthorlist}

  \icmlaffiliation{polimi}{Department of Electronics, Information and Bioengineering, Politecnico di Milano, Milan, Italy}
  \icmlaffiliation{comp}{SynapX, Beijing, China}
  \icmlaffiliation{tsu}{Department of Automation, Tsinghua University, Beijing, China}
  \icmlaffiliation{zju}{State Key Laboratory of Industrial Control Technology, Zhejiang University, Hangzhou, China}
  \icmlaffiliation{cas}{Institute of Computing Technology, Chinese Academy of Sciences, Beijing, China}
  \icmlaffiliation{bit}{School of Automation, Beijing Institute of Technology, Beijing, China}

  \icmlcorrespondingauthor{Tao Chen}{shubing@synapa.ai}
  \icmlcorrespondingauthor{Jie Zhou}{jzhou@mail.tsinghua.edu.cn}

  \icmlkeywords{Machine Learning, ICML}

  \vskip 0.3in
]

\printAffiliationsAndNotice{\textsuperscript{*}Equal contribution. \textsuperscript{\textdagger}Project leader.}  % no special notice (required even if empty)

\begin{abstract}
    Surface electromyography (sEMG) enables continuous hand-pose
    estimation, but learning representations that generalize
    across users and movements remains challenging.
    We introduce InterBand-EMG, which combines band-specific
    quantization with cross-band reconstruction and structured
    masked code prediction to learn transferable EMG
    representations for hand-pose tracking.
    We also introduce SynapMyo, an eight-channel EMG-to-pose
    dataset comprising 1,227.74 hand-hours from 499 users
    across 68 movement categories.
    Experiments on emg2pose and SynapMyo show that InterBand-EMG
    outperforms state-of-the-art baselines in joint-angle accuracy
    across all three generalization settings on both datasets. The source code and the SynapMyo dataset will be publicly
released upon acceptance of this paper.
\end{abstract}

\section{Introduction}
\label{sec:introduction}

Continuous hand-pose estimation recovers the joint-angle trajectories
that describe finger articulation during hand movements. It supports
applications such as demonstration capture, teleoperation, and human--robot interaction. Vision-based approaches provide accurate hand pose estimates~\citep{banerjee2025hot3d,han2022umetrack},
but their availability can be limited by objects occlusion and movements outside the camera's field of view~\citep{salter2024emg2pose}.
Surface electromyography (sEMG) offers a complementary sensing modality: skin-mounted electrodes record electrical activity associated with muscle activation, providing information for movement decoding without requiring the hand to remain visible.

Learning a transferable mapping from sEMG to hand pose is challenging
because anatomy, electrode placement, and sensor contact vary across
participants and recordings. The emg2pose benchmark provides approximately 370 hours of paired 16-channel sEMG and hand-pose data from 193 participants, with separate evaluations on held-out users,
movement stages, and both~\citep{salter2024emg2pose}. This benchmark
establishes a basis for studying generalization in continuous pose
decoding. Alongside task-supervised learning, an important question
is how pretraining on EMG without pose labels can produce
representations that improve subsequent pose estimation under
these generalization conditions.

Self-supervised biosignal learning addresses this question through
a range of prediction targets and training objectives. Some
approaches first discretize continuous signals using
vector-quantized (VQ) autoencoders, then learn representations by
predicting masked or future codes~\citep{jiang2024large,
cui2026brainrvq,barmpas2025neurorvq}. Others learn continuous
representations through signal reconstruction, latent prediction,
or contrastive objectives, without a separately pretrained discrete
tokenizer~\citep{yang2023biot,wang2024eegpt,wang2025cbramod}.
For continuous EMG-to-pose estimation, EMG-GPT predicts future codes
from a frozen tokenizer and transfers the resulting representations
to pose decoding~\citep{magni2026emg}. Its locally matched tracking
evaluation reports higher mean angular errors than the supervised
vEMG2Pose baseline across all three generalization splits, although
the models retain different input-processing pipelines. These
findings motivate further investigation of how EMG pretraining
targets influence downstream pose estimation.

\begin{figure*}[t]
    \centering
    \includegraphics[width=\textwidth]{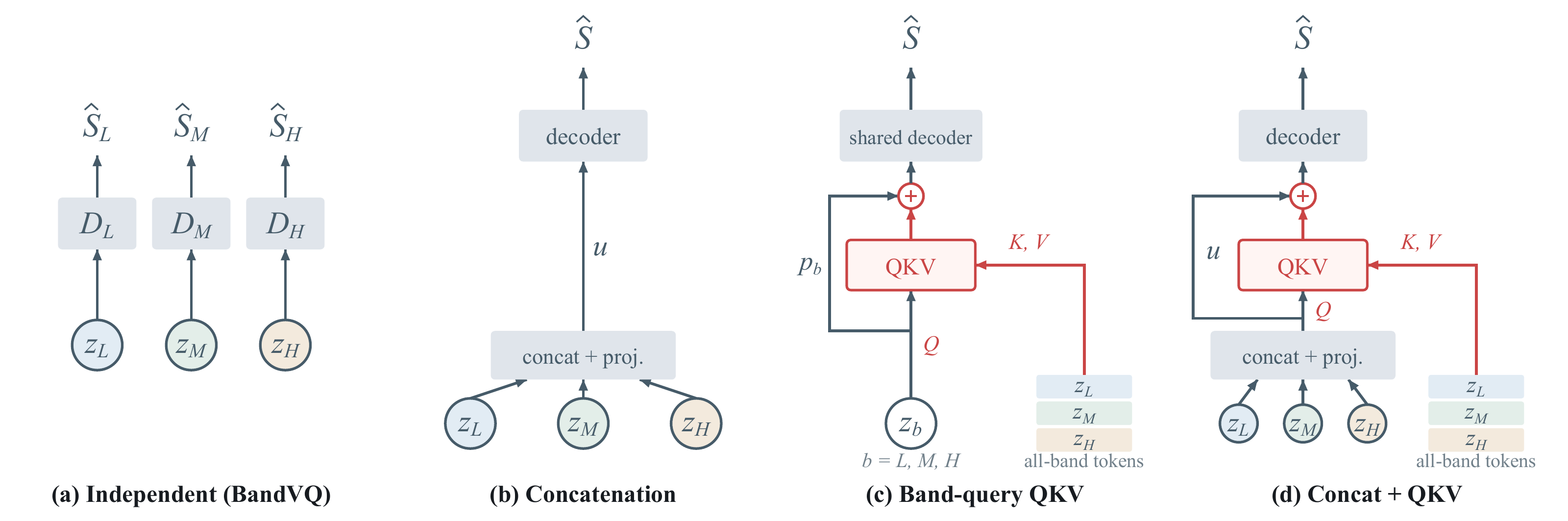}
    \caption{
    Reconstruction strategies for band-specific EMG tokenization.
    $z_L$, $z_M$, and $z_H$ denote quantized features from the low-,
    mid-, and high-frequency bands, respectively.
    (a) Independent reconstruction uses a separate decoder for each band
    (b) Concatenation combines and projects features from all three bands into a joint representation for full-spectrum reconstruction.
    (c) Band-query attention, used in InterBand-EMG, takes each band's features as queries and retrieves context from all bands. The retrieved context is added to that band's projected features and passed to a shared decoder with band-specific reconstruction heads.
    (d) Concatenation followed by attention uses the joint
    representation as both the query and residual input.
    In both attention variants, keys and values span all bands,
    electrodes, and temporal patches.
}
    \label{fig:reconstruction_architectures}
\end{figure*}

Frequency-aware tokenization offers one way to structure these
targets.  Surface EMG reflects the superposition of action potentials
from multiple motor units, with spectral characteristics influenced
by muscle properties, tissue filtering, and electrode
configuration~\citep{muceli2024tutorial}. Allocating separate
codebooks to different frequency bands may help retain signal
features that a single broadband vocabulary underrepresents.
BandVQ follows this principle for EEG by training an independent
VQ-VAE tokenizer for each frequency band and using the resulting
codes for shared-encoder pretraining~\citep{sukhbaatar2026bandvq}. 
However, its band tokenizers are learned independently: each
reconstructs its own band without access to features from the
others. This leaves open whether interaction during tokenizer
training can produce more useful discrete targets while retaining
separate vocabularies for individual bands.

We investigate this question through \textbf{InterBand-EMG}, a
self-supervised pretraining framework that combines band-specific
quantization with cross-band reconstruction. Our central hypothesis
is that coupling reconstruction across bands improves the
usefulness of the resulting codes as targets for representation
learning. Three encoders with separate codebooks tokenize the low-,
mid-, and high-frequency components of EMG. We compare independent
reconstruction with three interaction strategies: feature
concatenation, band-query attention, and concatenation followed by
attention (Figure~\ref{fig:reconstruction_architectures}). Our main model uses each band's quantized features as queries to retrieve context from all bands before reconstruction. Reconstruction losses therefore couple the learning of the band-specific codes while each band retains its own discrete vocabulary.
After tokenizer training, we freeze its parameters and use its
band-specific code indices as prediction targets. A shared encoder
receives masked broadband EMG and learns to predict the codes of
all three bands at masked positions. The encoder's continuous
representations are subsequently fine-tuned for hand-pose
estimation. This two-stage design allows us to examine whether
cross-band interaction during target construction improves
downstream representation quality.

To support larger-scale pretraining and evaluation, we also
introduce \textbf{SynapMyo}, an eight-channel EMG-to-pose dataset
intended for public release. SynapMyo contains 81,564 single-hand
recordings, totaling approximately 1,228 hand-hours across
499 participants and 68 annotated gesture and movement categories.
The collection includes glove-based recordings acquired using
Waveletech EMG wristbands and Manus motion-capture gloves, alongside
recordings annotated through vision-based hand reconstruction.
Both annotation sources are mapped to a common 20-dimensional
UmeTrack joint-angle representation and temporally aligned with
EMG. Hand-hours count each hand separately. We establish fixed
evaluation splits for held-out participants, held-out movement
tasks, and both. Recordings from the same acquisition episode,
together with temporally overlapping episodes from the same
participant, remain within a single split.

Our contributions are as follows:
\begin{itemize}
    \item We propose InterBand-EMG, a self-supervised framework that learns band-specific EMG codebooks through cross-band reconstruction and uses their discrete codes to retrain 
    a broadband encoder for continuous hand pose estimation.
    
    \item We introduce SynapMyo, an eight-channel EMG-to-pose
    dataset comprising approximately 1,228 hours of hand data with a common joint-angle representation and fixed evaluation splits for held-out participants and movement tasks.
    
    \item We achieve state-of-the-art joint-angle accuracy on
    both emg2pose and SynapMyo across all three generalization
    settings, reducing MAE by 0.3--4.2\% and 16.2--24.2\%,
    respectively, relative to the strongest evaluated baseline
    on each dataset.
\end{itemize}

\section{Related Work}
\label{sec:related_work}

\paragraph{Physiological signal foundation models.}
Self-supervised biosignal pretraining provides the methodological
basis for our approach. LaBraM, BrainRVQ, and NeuroRVQ learn
discrete signal representations through vector quantization
and use their codes as targets for masked or autoregressive
pretraining~\citep{jiang2024large,cui2026brainrvq,barmpas2025neurorvq}.
Complementary approaches learn continuous representations through
contrastive objectives, masked reconstruction, or latent
prediction~\citep{yang2023biot,wang2024eegpt,wang2025cbramod,
elouahidi2025reve,li2026eeg}.
Beyond EEG, QualityFM and SIGMA-PPG explore quality-aware
self-distillation and consistency-regularized tokenization,
respectively~\citep{guo2025qualityfm,guo2026sigma}. Complementary signal-processing work uses time--frequency--chirprate
representations to resolve components with crossing instantaneous
frequencies~\citep{chen2026instantaneous}.
Most directly related to our formulation, BandVQ learns
independent frequency-band tokenizers before shared-encoder
pretraining~\citep{sukhbaatar2026bandvq}.
Building on these ideas, we investigate whether cross-band
interaction during tokenizer reconstruction improves the
discrete targets used for EMG representation learning.

\paragraph{EMG-to-pose estimation and representation learning.}
NeuroPose and SensingDynamics demonstrated continuous
hand-kinematic decoding from wearable sEMG~\citep{liu2021neuropose,simpetru2022sensing}.
The emg2pose benchmark expanded evaluation to approximately
370 hand-hours from 193 participants and established
held-out-user, held-out-stage, and joint held-out conditions.
It also distinguishes pose regression from tracking with
a provided initial pose~\citep{salter2024emg2pose}.
Subsequent work shows that decoder parameterization and
training stability substantially affect position-versus-velocity
comparisons~\citep{hadidi2026reevaluating}, while NHN introduces
physiological inductive biases into task-supervised
decoding~\citep{wang2026neuromotor}.

Recent studies also investigate pretrained EMG representations.
TinyMyo uses self-supervised masked reconstruction and adapts
a shared encoder to multiple downstream tasks, including
hand kinematic regression~\citep{fasulo2026tinymyo}.
EMG-GPT predicts future codes from a frozen NeuroRVQ tokenizer
through a causal temporal model and transfers its representations
to pose regression and tracking~\citep{magni2026emg}.
A separate line of work uses paired kinematic information
during representation learning.
EMBridge aligns EMG and pose representations to support
generalization to unseen gesture classes~\citep{cui2026embridge},
whereas KinEMbed uses cross-modal contrastive alignment for
continuous hand-kinematic regression~\citep{gilardini2026kinembed}.
InterBand-EMG uses EMG-derived discrete targets without pose
labels during tokenizer training or encoder pretraining;
pose supervision is introduced during downstream fine-tuning.
Its methodological focus is the effect of cross-band
reconstruction on the targets used for masked pretraining.

\paragraph{Attention-based feature interaction.}
Hybrid CNN--Transformer models combine local feature extraction
with long-range temporal modeling for remote heart-rate and
respiratory-rate estimation~\citep{guo2023remote,chen2024actnet}.
Cross-attention allows queries from one feature stream to
retrieve information from another through key matching and
value aggregation~\citep{vaswani2017attention}.
MulT uses directional cross-attention between unaligned
modality sequences~\citep{tsai2019multimodal}, while CrossViT
exchanges information between image branches through
class-token queries~\citep{chen2021crossvit}.
Multimodal Bottleneck Transformers constrain information
exchange through a small set of shared latent
tokens~\citep{nagrani2021attention}.
In clinical forecasting, Clinical Visual Memory uses
modality-aware latent-query cross-attention to fuse heterogeneous
clinical inputs into a shared visual-token
memory~\citep{guo2026optical}.
For multivariate time series, Crossformer uses learned routers
to aggregate and redistribute information across
variables~\citep{zhang2023crossformer}, and TimeXer uses a
target series' global token to query representations of
external variables~\citep{wang2024timexer}.
These methods establish attention-based interaction as a
general architectural mechanism.
In InterBand-EMG, band-specific quantized features query
tokens from all frequency bands during reconstruction.
We study this interaction as a mechanism for jointly learning
separate band vocabularies, rather than as a new attention
operator.

\section{Methods}
\subsection{SynapMyo benchmark}

% Add \usepackage{graphicx} to the preamble.
\begin{figure*}[t]
    \centering
    \includegraphics[width=1.0\textwidth]{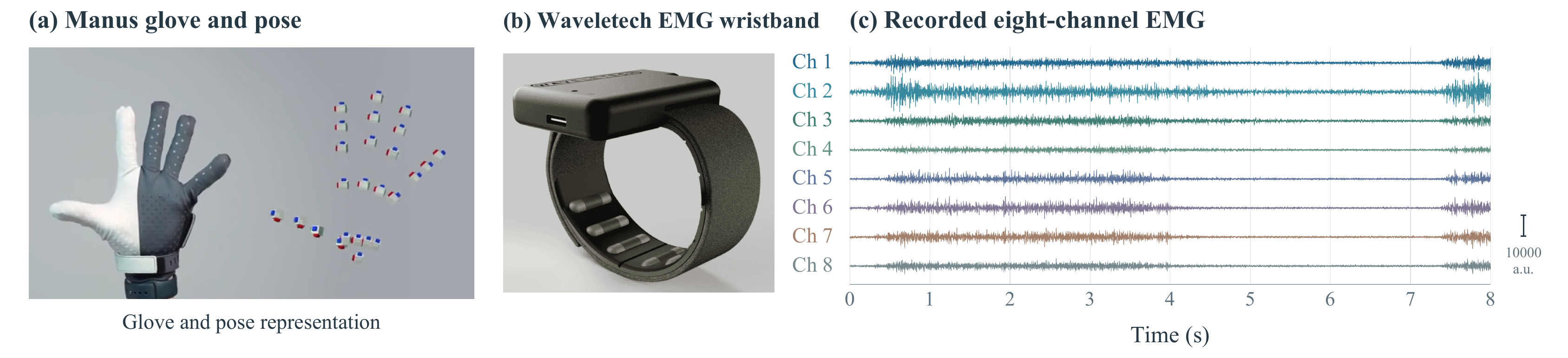}
    \caption{
    Glove-based motion capture and eight-channel EMG in SynapMyo.
    (a) A Manus glove shown alongside its hand-pose representation.
    (b) The Waveletech EMG wristband used for signal acquisition.
    (c) An eight-second excerpt of eight-channel EMG recorded with
    the wristband shown in (b) during right index-finger flexion
    and extension, sampled at $2\,\mathrm{kHz}$.
    The traces show the signals as stored in the dataset, without
    additional processing for visualization, and share a common
    amplitude scale, with vertical offsets for clarity.
    Amplitudes are shown in arbitrary units.
}
    \label{fig:synapmyo_acquisition}
\end{figure*}

\paragraph{Data acquisition.}
For glove-based recordings, participants perform hand gestures
while wearing eight-channel Waveletech EMG wristbands together
with Manus motion-capture gloves
(Figure~\ref{fig:synapmyo_acquisition}).
The wristbands record EMG at a nominal sampling rate of
$2\,\mathrm{kHz}$, while the gloves provide finger-kinematic
measurements at a manufacturer-specified nominal output rate
of $120\,\mathrm{Hz}$.
Acquisition tasks include individual-finger movements, predefined
hand gestures, continuous transitions between gestures,
unconstrained movements, and a smaller set of functional
activities, including writing, typing, and object manipulation
(Appendix~\ref{app:synapmyo_movements}).

\paragraph{Pose annotation and temporal alignment.}
We express both glove-derived and vision-derived annotations
in a common 20-dimensional UmeTrack joint-angle
representation~\citep{han2022umetrack}, with angles stored
in radians. For glove-based recordings, pose labels are
generated from Manus ergonomic angles using a mapping
calibrated separately for the left and right hands.
Calibration targets are obtained offline by fitting the
UmeTrack hand model to Manus skeletal landmarks.
The calibrated mapping produces 20-dimensional joint-angle
labels constrained to predefined joint limits.
Thumb MCP flexion is mapped affinely from its corresponding
Manus ergonomic channel.

% I check whith college, fixed is the old version

For vision-based recordings, we obtain three-dimensional
hand keypoints using HaMeR~\citep{pavlakos2024reconstructing}
and retain frames marked as valid by the upstream quality
filter. After accounting for handedness and aligning the
keypoints with the UmeTrack hand model, we recover joint
angles through kinematic retargeting~\citep{qin2023anyteleop}.

We place both annotation sources on the corresponding hand's EMG
time grid using recorded timestamps and linear interpolation.
The conversion applies timestamp-order corrections and extends
endpoint pose values outside the available pose interval;
no maximum interpolation gap is imposed.
These operations establish a common time grid but do not
constitute a measurement of hardware synchronization accuracy.
During preprocessing, the first 4,000 samples are removed from
EMG and all aligned fields together.
Appendix~\ref{app:synapmyo_qc} details timestamp handling,
interpolation, filtering, and audit coverage.

\begin{table}[t]
    \centering
    \caption{
        SynapMyo composition by pose annotation source.
        Durations are computed from retained samples at
        $2\,\mathrm{kHz}$, counting each hand separately.
        Users and movement categories overlap across sources.
        Valid label pass is the fraction of source hand frames
        accepted before interpolation; Total pools the accepted
        and total frame counts across sources.
    }
    \label{tab:synapmyo_composition}
    \small
    \setlength{\tabcolsep}{3pt}
    \renewcommand{\arraystretch}{1.08}
    \begin{tabularx}{\columnwidth}{@{}Xrrr@{}}
        \toprule
        Statistic & Vision & Glove & Total \\
        \midrule
        Recordings & 40,863 & 40,701 & 81,564 \\
        Hand-hours & 693.27 & 534.46 & 1,227.74 \\
        Users & 306 & 213 & 499 \\
        Movement categories & 59 & 66 & 68 \\
        Valid label pass (\%) & 74.98 & 100.00 & 93.74 \\
        \bottomrule
    \end{tabularx}
\end{table}

\paragraph{Dataset composition and quality control.}
SynapMyo contains 81,564 single-hand recordings, totaling approximately
1,228 hand-hours across 499 users and 68 annotated
gesture and movement categories.
Table~\ref{tab:synapmyo_composition} summarizes the subsets with
vision-derived and glove-derived pose annotations.
Movement categories are derived from recording-level task
annotations by normalizing aliases and separating explicitly
listed components of compound labels. A recording may therefore
contribute to multiple categories; these annotations do not
provide frame-level movement segmentation.

A recording is a single-hand sequence of EMG samples and
aligned joint-angle annotations stored in one HDF5 file.
A session corresponds to one source acquisition episode
and may include recordings from both hands. A recording group combines all recordings from the same episode and merges episodes from the same participant whose
time intervals overlap, including through transitive overlaps.
The benchmark contains 40,787 sessions and 40,756 recording groups. Durations are calculated from retained EMG samples at the nominal sampling rate of $2,\mathrm{kHz}$, counting each hand separately.

Quality control includes conversion-format checks,
file-integrity checks, sampled numerical audits, and
metadata screening. EMG samples must contain eight channels
and a timestamp, and glove messages must contain the required
20 ergonomic angle entries. A hand recording is exported only
when the required EMG and pose streams are available.
Corrupted exports and recordings too short for preprocessing
are excluded. Metadata screening additionally removes records
with designated source-specific exclusion tags, missing or
non-action task labels, or conflicting task assignments within
a recording group. Retained EMG undergoes causal high-pass
and power-line band-stop filtering, followed by removal of
the initial 4,000 samples from all aligned fields.
These checks establish structural consistency but do not
by themselves establish pose-label accuracy.
Detailed processing and audit coverage, the movement inventory,
and pose-label validation are described in
Appendices~\ref{app:synapmyo_qc},
\ref{app:synapmyo_movements}, and
\ref{app:synapmyo_label_quality}, respectively.

\begin{figure*}[t]
    \centering
    \includegraphics[width=\textwidth]{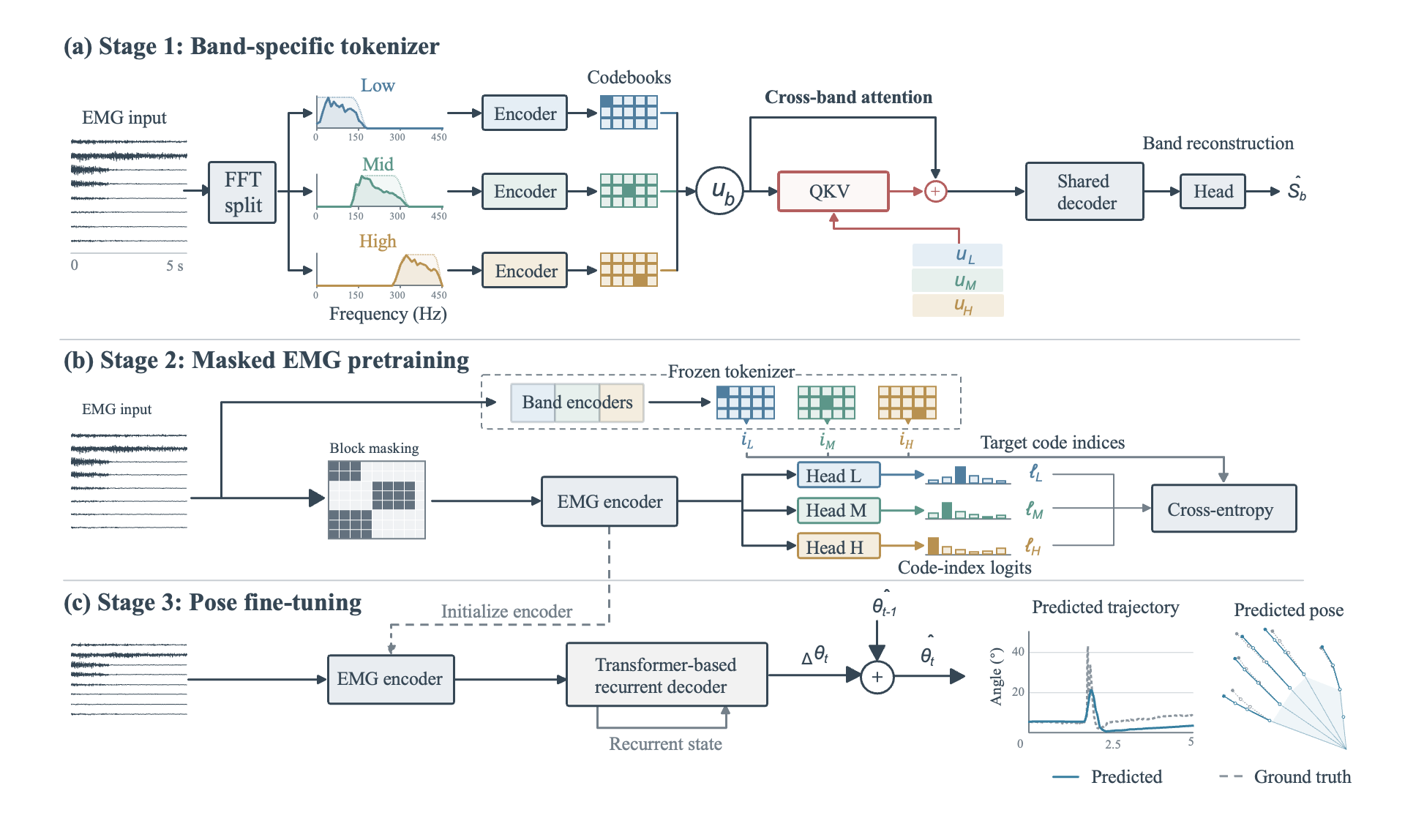}
    \caption{
    Overview of InterBand-EMG.
    (a) Band-specific tokenization. EMG is decomposed into three
    frequency bands and quantized using separate encoders and
    codebooks. Cross-band attention enables each band to retrieve
    information from all bands for spectral reconstruction.
    (b) Masked EMG pretraining. The frozen tokenizer produces
    band-specific target indices from unmasked EMG. A shared
    encoder processes masked EMG, and three classification heads
    predict the corresponding code indices using cross-entropy.
    (c) Pose fine-tuning. The pretrained encoder and a
    Transformer-based recurrent decoder are jointly fine-tuned
    to predict joint-angle increments, which are accumulated
    from a given initial pose. In the pose visualizations,
    solid blue and dashed gray denote predictions and
    ground truth, respectively.
}
\label{fig:interband_overview}
\end{figure*}

\paragraph{Prediction task and evaluation metrics.}
We evaluate initial-pose-conditioned hand tracking from
eight-channel EMG, following the task formulation of
emg2pose~\citep{salter2024emg2pose}.
Each example comprises 0.895 seconds of preceding EMG context
and a 5-second prediction interval.
The model receives the 20-dimensional
reference pose at the start of this interval
and predicts the remaining trajectory without further
pose observations. Model state is reset for each window, and context is excluded from scoring.

The decoder produces 250 pose estimates per window at a nominal rate of $50,\mathrm{Hz}$. We report mean absolute angular error over all 20 angles in degrees after interpolation to the $2,\mathrm{kHz}$ reference grid. We additionally report mean Euclidean distance over 17 UmeTrack landmarks in millimeters, evaluated at $50,\mathrm{Hz}$ under a fixed hand geometry.
For each generalization condition, errors are first averaged
over each user's valid evaluated frames. We then report the
equally weighted mean and sample standard deviation across users.
Window construction, temporal resampling, landmark selection,
and checkpoint selection are specified in Appendix~\ref{app:synapmyo_evaluation}.

\paragraph{Generalization settings.}
We evaluate three generalization conditions using a shared
training partition: \textbf{User} tests held-out participants
performing tasks represented in training; \textbf{Task} tests
training participants performing held-out annotated acquisition
tasks; and \textbf{User+Task} tests held-out participants
performing held-out tasks.
The 499 participants are partitioned into 400 training,
49 validation, and 50 test participants. Task-only validation
and test subsets contain participants from the training
participant pool.

Eight annotated movement categories are reserved for held-out-task evaluation and excluded from training.
 Any recording whose normalized task annotation contains
at least one reserved category, including within a compound
annotation, is assigned to a held-out-task subset.
Validation and test share these eight categories,
listed in Appendix~\ref{app:synapmyo_movements}, but use disjoint
recording groups. No recording group is shared between
training, validation, or test, or evaluation subsets.
Table~\ref{tab:synapmyo_generalization} reports the partition
sizes. The shared training partition contains 57,439 recordings,
totaling 856.25 hand-hours.

\begin{table}[t]
    \centering
    \caption{Statistics of the SynapMyo dataset splits.
             Users and Records denote the numbers of distinct users
             and single-hand recordings in each row, respectively.
             Durations are reported in hand-hours.}
    \label{tab:synapmyo_generalization}
    \small
    \setlength{\tabcolsep}{2.5pt}
    \renewcommand{\arraystretch}{1.08}
    \begin{tabularx}{\columnwidth}{@{}lXrrr@{}}
        \toprule
        Split & Setting & Users & Records & Hours \\
        \midrule
        Train & Shared & 400 & 57,439 & 856.25 \\
        \midrule
        Val. & User & 49 & 4,938 & 81.68 \\
        Val. & Task & 394 & 6,106 & 85.93 \\
        Val. & User+Task & 49 & 1,074 & 16.97 \\
        Val. & All & 443 & 12,118 & 184.58 \\
        \midrule
        Test & User & 50 & 4,758 & 82.47 \\
        Test & Task & 395 & 6,152 & 86.25 \\
        Test & User+Task & 50 & 1,097 & 18.19 \\
        Test & All & 445 & 12,007 & 186.91 \\
        \midrule
        All & All & 499 & 81,564 & 1,227.74 \\
        \bottomrule
    \end{tabularx}
\end{table}

\subsection{InterBand-EMG}
InterBand-EMG employs a three-stage training pipeline
(Figure~\ref{fig:interband_overview}). First, we learn a
band-specific tokenizer through cross-band reconstruction. Then, we use its discrete code indices to supervise masked EMG pretraining. Finally, we fine-tune the pretrained encoder for continuous hand-pose tracking.

\paragraph{Band-specific tokenization with cross-band reconstruction}
\label{sec:band_tokenization}

We divide the preprocessed EMG signal into non-overlapping temporal patches
and apply three overlapping frequency-domain windows, indexed
$\mathcal{B}=\{L,M,H\}$, with nominal ranges of
0--150\,Hz, 150--300\,Hz, and 300--450\,Hz. %you apllied a high pass filter >20Hz the first band 20-150???
For channel $c$ and patch $p$, we obtain the band-specific
time-domain signal as
\begin{equation}
\begin{split}
&\mathbf X_{c,p}=\mathcal F(\mathbf x_{c,p}),
\qquad 
X^{(b)}_{c,p}[k]=w_b[k]\,X_{c,p}[k],
\qquad
b\in\mathcal{B},
\\
&\quad
\mathbf{x}^{(b)}_{c,p}=\mathcal F^{-1}
\!\left[X^{(b)}_{c,p}\right]
\end{split}
\end{equation}
where $\mathcal{F}$ denotes the Fourier transform, $k$ indexes frequency bins, and $w_b$ is a fixed, smoothly tapered frequency-domain window. For the full Fourier spectrum, each window is symmetric across positive and negative frequencies, ensuring real-valued inverse transforms.

The adjacent bands overlap at 125--175\,Hz and 275--325\,Hz, respectively. Within each overlap region, complementary sine and cosine weights transfer emphasis smoothly from the lower to the higher band. This allows both branches to retain information near the nominal boundary. The high-frequency window additionally tapers from 420\,Hz to zero at 450\,Hz.
Three independent encoders then process the resulting band-specific time-domain
patches.

Each encoder produces normalized feature vectors
$\mathbf{h}_{b,j}$, where $b$ identifies the frequency band
and $j$ indexes a channel--patch position.
For each band, a separate codebook
$\mathcal{C}_b=\{\mathbf{e}_{b,l}\}_{l=0}^{L_b-1}$
contains $L_b=1024$ normalized code vectors.
We assign each feature to its nearest code vector:
\begin{equation}
    i_{b,j}
    =
    \operatorname*{arg\,min}_{l\in\{0,\dots,L_b\}}
    \left\|
        \mathbf{h}_{b,j}-\mathbf{e}_{b,l}
    \right\|_2^2,
    \qquad
    \mathbf{z}_{b,j}=\mathbf{e}_{b,i_{b,j}}.
\end{equation}
Here, $i_{b,j}$ is the discrete code index and
$\mathbf{z}_{b,j}$ is the corresponding quantized feature
vector. Each channel--patch position therefore receives
one index and one quantized vector per band.

Cross-band interaction operates on the quantized vectors.
For each band, we stack these vectors into a token matrix
$\mathbf{Z}_b$, project them to a common feature dimension,
and add a learned band embedding to obtain $\mathbf{U}_b$.
Adding channel and temporal embeddings $\mathbf{P}$ gives
the position-aware tokens. We concatenate these tokens
across bands to construct the shared memory:
\begin{equation}
\mathbf{M}
=
\operatorname{Concat}_{\mathrm{tokens}}
\bigl(
\mathbf{U}_L+\mathbf{P},
\mathbf{U}_M+\mathbf{P},
\mathbf{U}_H+\mathbf{P}
\bigr).
\end{equation}
Each band supplies its own queries, while keys and values
span all bands, channels, and temporal patches.

For attention head $h$,
\begin{equation}
\begin{aligned}
    \mathbf{Q}_b^h &= (\mathbf{U}_b+\mathbf{P})\mathbf{W}_Q^h,
    \quad
    \mathbf{K}^h = \mathbf{M}\mathbf{W}_K^h,
    \quad
    \mathbf{V}^h = \mathbf{M}\mathbf{W}_V^h, \\
    \mathbf{R}_b^h
    &=
    \operatorname{softmax}
    \left(
        \frac{\mathbf{Q}_b^h(\mathbf{K}^h)^\top}{\sqrt{d_h}}
    \right)\mathbf{V}^h, \\
    \widetilde{\mathbf{U}}_b
    &=
    \mathbf{U}_b+
    \operatorname{Concat}_{h}(\mathbf{R}_b^h)\mathbf{W}_O.
\end{aligned}
\end{equation}
where \(d_h\) is the query/key dimension of head \(h\) and softmax operates over memory tokens.
The residual connection retains each band's own features
while incorporating information retrieved from the shared
memory. Figure~\ref{fig:reconstruction_architectures} also
illustrates feature-wise concatenation and concatenation
followed by attention as alternative interaction mechanisms.

A shared Transformer decoder processes the updated features,
followed by separate band-specific heads for amplitude and
phase reconstruction. We combine the predictions within
each output type using fixed spectral memberships:
% in the figure you name S and not Y the spectrum!
\begin{equation}
\widehat{S}_{j}^{(o)}(k)
=
\sum_{b\in\mathcal{B}}
m_b[k]\ \widehat{S}_{b,j}^{(o)}[k],
\qquad
m_b[k]=w_b[k]^2,
\end{equation}
where $o\in\{\mathrm{amp},\mathrm{phase}\}$ identifies the
output type, $j$ indexes a channel--patch position, and
$k$ indexes a frequency bin.
Here, $\widehat{Y}_{b,j}^{(o)}[k]$ is the prediction of
band $b$'s corresponding reconstruction head at bin $k$.
The targets are the magnitude and principal phase of the
full FFT of the original broadband patches. Magnitude and
phase targets are standardized separately within each
example, using statistics pooled over channels, patches,
and frequency bins.

The amplitude objective $\mathcal{L}_A$ uses squared
reconstruction errors normalized by fixed per-frequency
scales estimated from a training subset. Within each band,
errors are averaged over frequency bins using the weights
$m_b[k]$; the resulting three band losses are then averaged
with equal weight. The phase objective $\mathcal{L}_{\phi}$
uses squared error between predicted and target standardized
principal-phase values, weighted by
$r[k]=\sum_{b\in\mathcal{B}}m_b[k]$
over the retained frequency bins.
The tokenizer objective is
\begin{equation}
    \mathcal{L}_{\mathrm{tok}}
    =
    \mathcal{L}_A
    + \mathcal{L}_{\phi}
    + \beta\mathcal{L}_{\mathrm{commit}},
\end{equation}
where $\beta=1$. The commitment loss penalizes the discrepancy between normalized encoder features and their assigned code vectors, with gradients stopped through the code vectors.
Codebooks are updated using exponential moving averages.
Cross-band attention allows reconstruction losses to influence features from multiple bands while retaining separate band-specific codebooks. Appendix~\ref{app:band_information} analyzes the predictive
contributions of these frequency ranges through band-combination,
band-removal, and temporal-shift experiments.

\paragraph{Structured Masked Code Prediction}
\label{sec:masked_code_prediction}

We freeze the trained tokenizer and apply it to unmasked
EMG to obtain three band-specific target indices at each
channel--patch position. A shared encoder is then pretrained
to predict these indices from masked broadband EMG. We construct a mask $\mathcal{M}$ on the channel--time patch
grid by combining four patterns: temporal spans across
multiple channels, local channel--time blocks, long temporal
spans within small channel subsets, and entire channels.
The mask covers 50\% of patch positions. A second view uses
the complementary mask $\overline{\mathcal{M}}$, so each
position is masked in one view and visible in the other.
Both views use the same encoder and prediction heads.

For a view with masked positions $\mathcal{V}$, the encoder
receives visible EMG patches and learned mask tokens replacingthe masked patch representations. Three classification heads predict the low-, mid-, and high-frequency code indices at masked positions, each producing logits over its corresponding 1,024-entry codebook. The pretraining objective is
\begin{equation}
\begin{aligned}
    \mathcal{L}_{\mathrm{pre}}
    &=
    \frac{1}{3}\sum_{b\in\mathcal{B}}
    \left[
        \mathcal{L}_b(\mathcal{M})
        +
        \mathcal{L}_b(\overline{\mathcal{M}})
    \right], \\
    \mathcal{L}_b(\mathcal{V})
    &=
    \frac{1}{|\mathcal{V}|}
    \sum_{j\in\mathcal{V}}
    \operatorname{CE}
    \left(
        \boldsymbol{\ell}_{b,j}^{\mathcal{V}},
        i_{b,j}
    \right),
\end{aligned}
\end{equation}
where $\boldsymbol{\ell}_{b,j}^{\mathcal{V}}$ denotes the
classification logits for band $b$ at position $j$ in the
view masked by $\mathcal{V}$, and $i_{b,j}$ is the frozen
tokenizer's target index. Losses are averaged over masked
positions and bands, summed across the two views, and
averaged over training examples.

\paragraph{Pose Fine-Tuning}
\label{sec:pose_finetuning}

We jointly fine-tune the pretrained EMG encoder and a
Transformer-based recurrent decoder for initial-pose-conditioned hand tracking. The decoder receives encoder features and conditions on the previous predicted pose and angle increment while maintaining a recurrent hidden state.
At each prediction step, it outputs an angle increment,
which updates the pose as
\begin{equation}
    \widehat{\boldsymbol{\theta}}_t
    =
    \widehat{\boldsymbol{\theta}}_{t-1}
    + \Delta\widehat{\boldsymbol{\theta}}_t,
    \qquad
    \widehat{\boldsymbol{\theta}}_0
    = \boldsymbol{\theta}_0.
\end{equation}
where $\boldsymbol{\theta}_0$ is the supplied initial
reference pose. Subsequent poses are generated
autoregressively without additional reference-pose inputs.
Predictions are produced at a nominal rate of $50\,\mathrm{Hz}$ and linearly interpolated to the aligned $2\,\mathrm{kHz}$ annotation grid.

The fine-tuning objective is
\begin{equation}
    \mathcal{L}_{\mathrm{pose}}
    =
    \mathcal{L}_{\mathrm{angle}}
    + 0.01\,\mathcal{L}_{\mathrm{tip}}.
\end{equation}
Here, $\mathcal{L}_{\mathrm{angle}}$ is the mean absolute
error over all 20 joint angles on the aligned annotation
grid, with angles expressed in radians. The fingertip loss $\mathcal{L}_{\mathrm{tip}}$ is the mean
Euclidean distance over the five fingertips, expressed in
millimeters and computed at $50,\mathrm{Hz}$.
Predicted and reference fingertip positions are obtained
by applying the same fixed UmeTrack hand model to their
respective joint-angle trajectories.
%The coefficient 0.01 depends on the stated units—radians and millimeters. It was fixed beforehand or selected using validation data?

\providecommand{\emgscore}[2]{%
    \ensuremath{#1\,{\scriptstyle\pm#2}}%
}

% Requires \usepackage{booktabs}.
\providecommand{\emgscore}[2]{%
    \ensuremath{#1\,{\scriptstyle\pm#2}}}
\providecommand{\bestemgscore}[2]{%
    \ensuremath{\mathbf{#1}\,{\scriptstyle\pm\mathbf{#2}}}}

\begin{table*}[h]
\centering
\caption{Tracking performance on emg2pose dataset.
MAE is joint-angle mean absolute error ($^\circ$); LD and FD are
landmark and five-fingertip distances (mm).
Lower is better; bold marks the lowest reported mean,
and dashes indicate unavailable results.
Published MAE/LD are user means $\pm$ between-user SD from
Table~1(b) of reference~$d$, except EMG-GPT from reference~$c$.
FD for Static, emg2pose, vemg2pose, and EMG-GPT is reevaluated
by us with the same user aggregation.
NeuroPose and Sensing are our initial-pose-conditioned
Tracking adaptations trained from scratch, reporting frame
means $\pm$ frame SD.
Ours reports the mean $\pm$ sample SD across three training seeds,
using frame-averaged metrics within each run.
Reevaluated spatial errors use our hand template, which differs
from that used for published LD; EMG-GPT retains its native
50\,Hz evaluation windows.
Pos/Vel denote position/velocity decoding; ST/MT denote
single-task/multi-task training.
References:
$^{a}$\citet{hadidi2026reevaluating};
$^{b}$\citet{salter2024emg2pose};
$^{c}$\citet{magni2026emg};
$^{d}$\citet{wang2026neuromotor};
$^{e}$\citet{liu2021neuropose};
$^{f}$\citet{simpetru2022sensing}.}
\label{tab:emg2pose_main}

\footnotesize
\setlength{\tabcolsep}{2pt}
\renewcommand{\arraystretch}{1.12}

\begin{tabular*}{\textwidth}
    {@{\extracolsep{\fill}}l*{9}{c}@{}}
    \toprule
    & \multicolumn{3}{c}{User}
    & \multicolumn{3}{c}{Stage}
    & \multicolumn{3}{c}{User+Stage} \\
    \cmidrule(lr){2-4}
    \cmidrule(lr){5-7}
    \cmidrule(l){8-10}
    Method
    & MAE ($^\circ$) & LD (mm) & FD (mm)
    & MAE ($^\circ$) & LD (mm) & FD (mm)
    & MAE ($^\circ$) & LD (mm) & FD (mm) \\
    \midrule

    Static$^{a}$
    & \emgscore{10.48}{1.81}
    & \emgscore{15.71}{2.98}
    & \emgscore{23.27}{4.47}
    & \emgscore{14.86}{1.91}
    & \emgscore{21.81}{2.58}
    & \emgscore{33.11}{3.96}
    & \emgscore{14.29}{1.86}
    & \emgscore{21.48}{2.33}
    & \emgscore{32.55}{3.44} \\

    emg2pose$^{b}$
    & \emgscore{11.33}{1.10}
    & \emgscore{14.75}{1.53}
    & \emgscore{21.74}{2.31}
    & \emgscore{15.39}{1.50}
    & \emgscore{20.88}{1.92}
    & \emgscore{31.87}{3.08}
    & \emgscore{15.05}{1.22}
    & \emgscore{20.69}{1.39}
    & \emgscore{31.55}{2.20} \\

    NeuroPose$^{e}$
    & \emgscore{9.33}{7.24}
    & \emgscore{12.36}{10.26}
    & \emgscore{19.55}{16.70}
    & \emgscore{13.75}{8.51}
    & \emgscore{18.11}{11.11}
    & \emgscore{29.50}{18.53}
    & \emgscore{13.17}{8.27}
    & \emgscore{17.78}{10.97}
    & \emgscore{28.90}{18.32} \\

    Sensing$^{f}$
    & \emgscore{10.24}{8.75}
    & \emgscore{14.26}{12.93}
    & \emgscore{22.59}{20.89}
    & \emgscore{14.74}{9.78}
    & \emgscore{20.14}{13.36}
    & \emgscore{32.74}{21.90}
    & \emgscore{14.14}{9.59}
    & \emgscore{19.88}{13.33}
    & \emgscore{32.23}{21.82} \\

    EMG-GPT$^{c}$
    & \emgscore{8.89}{0.93}
    & \emgscore{11.59}{1.24}
    & \emgscore{17.10}{1.83}
    & \emgscore{12.42}{1.43}
    & \emgscore{16.88}{1.80}
    & \emgscore{25.48}{2.80}
    & \emgscore{12.18}{1.25}
    & \emgscore{16.93}{1.36}
    & \emgscore{25.53}{2.05} \\

    vemg2pose$^{b}$
    & \emgscore{7.70}{0.98}
    & \emgscore{10.22}{1.54}
    & \emgscore{15.04}{2.35}
    & \emgscore{11.20}{1.43}
    & \emgscore{15.18}{1.81}
    & \emgscore{22.92}{2.85}
    & \emgscore{11.03}{1.01}
    & \emgscore{15.35}{1.46}
    & \emgscore{23.13}{2.25} \\

    Pos-ST$^{a}$
    & \emgscore{7.35}{0.92} & \emgscore{9.67}{1.48} & --
    & \emgscore{10.21}{1.32} & \emgscore{13.73}{1.69} & --
    & \emgscore{10.25}{0.93} & \emgscore{14.18}{1.35} & -- \\

    Vel-ST$^{a}$
    & \emgscore{7.64}{0.94} & \emgscore{10.02}{1.52} & --
    & \emgscore{10.76}{1.34} & \emgscore{14.43}{1.71} & --
    & \emgscore{10.75}{0.93} & \emgscore{14.82}{1.40} & -- \\

    Pos-MT$^{a}$
    & \emgscore{7.36}{0.90} & \emgscore{9.68}{1.46} & --
    & \emgscore{10.28}{1.33} & \emgscore{13.81}{1.70} & --
    & \emgscore{10.29}{0.97} & \emgscore{14.22}{1.37} & -- \\

    Vel-MT$^{a}$
    & \emgscore{7.64}{0.95} & \emgscore{10.00}{1.53} & --
    & \emgscore{10.77}{1.36} & \emgscore{14.44}{1.75} & --
    & \emgscore{10.75}{0.97} & \emgscore{14.81}{1.41} & -- \\

    NHN$^{d}$
    & \emgscore{7.30}{0.88} & \emgscore{9.55}{1.41} & --
    & \emgscore{10.12}{1.31} & \emgscore{13.59}{1.69} & --
    & \emgscore{10.16}{0.99} & \emgscore{14.03}{1.33} & -- \\

    \midrule
    \textbf{Ours}
    & \bestemgscore{7.28}{0.92}
    & \bestemgscore{8.81}{1.30}
    & \bestemgscore{13.79}{2.16}
    & \bestemgscore{9.70}{1.31}
    & \bestemgscore{11.95}{1.56}
    & \bestemgscore{19.19}{2.59}
    & \bestemgscore{10.00}{0.96}
    & \bestemgscore{12.73}{1.19}
    & \bestemgscore{20.42}{2.05} \\
    \bottomrule
\end{tabular*}
\end{table*}

\providecommand{\bestemgscore}[2]{\ensuremath{\mathbf{#1}\,{\scriptstyle\pm\mathbf{#2}}}}
\providecommand{\emgscore}[2]{\ensuremath{#1\,{\scriptstyle\pm#2}}}
% Requires \usepackage{booktabs} in the preamble.
\begin{table*}[!h]
\centering
    \caption{Tracking performance on SynapMyo.
    All methods use eight-channel EMG and a provided initial pose.
    MAE denotes joint-angle mean absolute error ($^\circ$);
    LD and FD denote mean distances over 17 landmarks and five fingertips,
    respectively (mm).
    Values are mean $\pm$ sample SD across users: errors are first averaged
    over each user's valid frames, then aggregated with equal weight per user.
    User, Task, and User+Task contain 50, 395, and 50 evaluated users, respectively.
    Static holds the initial pose fixed; NeuroPose and Sensing
    are initial-pose-conditioned Tracking adaptations trained from scratch.
    Lower is better, and bold marks the lowest mean in each column.
    References: $^{b}$\citet{salter2024emg2pose};
    $^{e}$\citet{liu2021neuropose};
    $^{f}$\citet{simpetru2022sensing}.}
    \label{tab:synapmyo_main}
    \footnotesize
    \setlength{\tabcolsep}{2pt}
    \renewcommand{\arraystretch}{1.12}
    \begin{tabular*}{\textwidth}
    {@{\extracolsep{\fill}}l*{9}{c}@{}}
    \toprule
    & \multicolumn{3}{c}{User}
    & \multicolumn{3}{c}{Task}
    & \multicolumn{3}{c}{User+Task} \\
    \cmidrule(lr){2-4}
    \cmidrule(lr){5-7}
    \cmidrule(l){8-10}
    Method
    & MAE ($^\circ$) & LD (mm) & FD (mm)
    & MAE ($^\circ$) & LD (mm) & FD (mm)
    & MAE ($^\circ$) & LD (mm) & FD (mm) \\
    \midrule
    Static
        & \emgscore{10.93}{1.35} & \emgscore{15.50}{3.26} & \emgscore{25.81}{5.23}
        & \emgscore{9.51}{2.95} & \emgscore{13.20}{4.28} & \emgscore{22.19}{7.21}
        & \emgscore{9.34}{2.91} & \emgscore{12.64}{3.96} & \emgscore{21.29}{6.76} \\
    
    emg2pose$^{b}$
        & \emgscore{14.03}{1.42} & \emgscore{20.19}{4.08} & \emgscore{33.36}{6.17}
        & \emgscore{13.43}{2.03} & \emgscore{19.57}{4.30} & \emgscore{32.10}{6.73}
        & \emgscore{13.12}{1.50} & \emgscore{18.59}{3.67} & \emgscore{30.59}{5.71} \\
    
    vemg2pose$^{b}$
        & \emgscore{6.04}{0.90} & \emgscore{8.31}{1.81} & \emgscore{13.67}{2.93}
        & \emgscore{6.91}{1.61} & \emgscore{9.69}{2.49} & \emgscore{16.21}{4.16}
        & \emgscore{6.86}{1.56} & \emgscore{9.41}{2.30} & \emgscore{15.82}{3.87} \\
    
    NeuroPose$^{e}$
        & \emgscore{9.11}{1.11} & \emgscore{11.85}{2.20} & \emgscore{19.50}{3.58}
        & \emgscore{9.15}{2.41} & \emgscore{12.40}{3.20} & \emgscore{20.88}{5.41}
        & \emgscore{8.92}{2.42} & \emgscore{11.90}{3.19} & \emgscore{20.08}{5.49} \\
    
    Sensing$^{f}$
        & \emgscore{10.56}{1.27} & \emgscore{14.73}{2.90} & \emgscore{24.58}{4.73}
        & \emgscore{9.45}{2.68} & \emgscore{13.17}{3.72} & \emgscore{22.17}{6.36}
        & \emgscore{9.26}{2.67} & \emgscore{12.63}{3.49} & \emgscore{21.29}{6.02} \\
    
    \midrule
    \textbf{Ours}
        & \bestemgscore{5.06}{0.88} & \bestemgscore{6.87}{1.57} & \bestemgscore{11.18}{2.54}
        & \bestemgscore{5.24}{1.25} & \bestemgscore{7.19}{1.75} & \bestemgscore{11.82}{2.93}
        & \bestemgscore{5.70}{1.34} & \bestemgscore{7.73}{1.92} & \bestemgscore{12.81}{3.16} \\
    \bottomrule
    \end{tabular*}
\end{table*}

\section{Experiments}
\label{sec:experiments}

\subsection{Experimental setup}
\label{sec:experimental_setup}
We evaluate InterBand-EMG on the emg2pose
benchmark~\citep{salter2024emg2pose} and SynapMyo using
their respective training, validation, and test partitions.
Both evaluations use initial-pose-conditioned hand tracking
under held-out-user, held-out-task, and combined conditions;
tasks correspond to movement stages in emg2pose.
The primary metric is joint-angle MAE in degrees, supplemented by landmark distance in millimeters.

For emg2pose, we report published tracking results for
vEMG2Pose~\citep{salter2024emg2pose},
EMG-GPT~\citep{magni2026emg}, and the Neuromotor Hierarchy
Network (NHN)~\citep{wang2026neuromotor}.
These entries are explicitly identified as published results
and retain their original evaluation protocols; they provide
context rather than a fully controlled comparison.
For SynapMyo, we train supervised baselines using the
publicly released emg2pose and vemg2pose
architectures, adapted to eight-channel input and evaluated
under the common SynapMyo protocol.
InterBand-EMG is trained separately on each dataset.
Tokenizer training and masked pretraining use only EMG
from the corresponding training partition, followed by
supervised fine-tuning on its paired EMG--pose recordings.
Model configurations, optimization settings, baseline
adaptations, and reporting conventions are provided in
Appendix~\ref{app:experimental_setup}.

\subsection{Comparison with baselines}
\label{sec:baseline_comparison}

Table~\ref{tab:emg2pose_main} reports tracking performance
on emg2pose under User, Stage, and User+Stage generalization.
Published results retain their source evaluation conventions
and serve as reference comparisons; locally evaluated results
are identified separately.
Differences in aggregation, temporal sampling, and hand
geometry are documented in
Appendix~\ref{app:published_baselines}.
Table~\ref{tab:synapmyo_main} compares InterBand-EMG with
Static, the supervised emg2pose and
vemg2pose architectures, and initial-pose-conditioned adaptations of NeuroPose and SensingDynamics on SynapMyo.
All methods use the common SynapMyo evaluation protocol
under User, Task, and User+Task generalization.
Both tables report joint-angle MAE, landmark distance,
and fingertip distance with a provided initial pose.
Appendix~\ref{app:synapmyo_movements} provides category-wise
results and annotation-source composition, while
Appendix~\ref{app:synapmyo_qualitative} presents selected
qualitative comparisons with all five baselines.
Supplementary analyses examine the contribution of pretraining
relative to training from scratch
(Appendix~\ref{app:pretrained_scratch}) and generalization
with reduced pose supervision
(Appendix~\ref{app:label_efficiency}).

\subsection{Ablation of Cross-Band Interaction}
\label{sec:interaction_ablation}

\begin{figure}[t]
    \centering
    \includegraphics[width=\columnwidth]{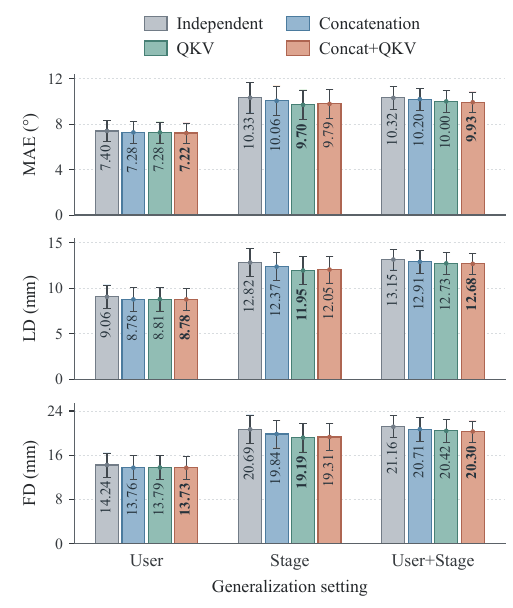}
    \caption{Reconstruction-design comparison on emg2pose Tracking.
    Panels report MAE ($^\circ$), LD (mm), and FD (mm).
    Bars and error bars show equal-user means and one sample
    standard deviation across users; lower is better.
    Bold values mark the lowest mean within each condition and metric.}
    \label{fig:interaction_ablation}
\end{figure}

Figure~\ref{fig:interaction_ablation} compares the four reconstruction
designs in Figure~\ref{fig:reconstruction_architectures}.
All three configurations with cross-band interaction achieve lower
MAE, LD, and FD than the Independent reference across the three
generalization conditions. Band-query attention (QKV) obtains
the lowest errors on Stage ($9.70^\circ$ MAE, $11.95$\,mm LD,
and $19.19$\,mm FD), whereas Concat+QKV performs best on User
and User+Stage across all three metrics. The relative performance
of band-specific and concatenated queries therefore varies across
generalization conditions.

\section{Discussion}

We introduced InterBand-EMG, a self-supervised framework
that combines band-specific quantization, cross-band
reconstruction, and masked code prediction for continuous
hand-pose tracking.
We also introduced SynapMyo, an eight-channel EMG-to-pose
dataset comprising 81,564 single-hand recordings,
totaling 1,227.74 hand-hours from 499 users across
68 movement categories.
SynapMyo provides a large-scale benchmark with fixed
evaluation splits for generalization to held-out users,
acquisition tasks, and their combination.
Experiments on SynapMyo and emg2pose demonstrate the
effectiveness of InterBand-EMG, while supplementary
analyses support the benefits of pretraining and
adaptation with reduced pose supervision.

Several limitations remain. SynapMyo has uneven movement coverage and heterogeneous
annotation sources, with limited dynamic information for
some glove-derived joint components.
Reliability information should accompany these derived annotations,
consistent with evidence of over-trust in downstream LLM use of
estimated physiological measurements~\citep{guo2026derived}.
Moreover, held-out acquisition tasks can resemble training
tasks, so these splits do not necessarily represent entirely
unseen movement primitives.
Our evaluation focuses on offline Tracking with a provided
initial pose; real-time latency, temporal stability, and
robustness to sensor repositioning remain unverified.
The reconstruction-design comparisons also include differences
in reconstruction objectives and masking configurations,
limiting attribution of the observed gains solely to the
interaction mechanism.

Future work will prioritize broader and more balanced task
coverage, improved annotation validation, and comparisons
that isolate cross-band interaction under matched training
conditions. We will also investigate how representation quality scales
with model capacity and unlabeled EMG data.
Extending evaluation to causal streaming inference, changing
sensor placements, and reduced dependence on initial-pose
information will help assess the framework's suitability
for practical wearable interaction and teleoperation.

\section{Conclusiom}

We introduced InterBand-EMG, a self-supervised framework
that combines band-specific quantization and cross-band
reconstruction for continuous
hand-pose tracking.
We also introduced SynapMyo, an eight-channel EMG-to-pose
dataset comprising 81,564 single-hand recordings,
totaling 1,227.74 hand-hours from 499 users across
68 movement categories.
Experiments on SynapMyo and emg2pose demonstrate the
effectiveness of InterBand-EMG, while supplementary
analyses support the benefits of pretraining and
adaptation with reduced pose supervision.

%\section*{Acknowledgements}
%\textbf{Do not} include acknowledgements in the initial version of the paper submitted for blind review.
%If a paper is accepted, the final camera-ready version can (and usually should)include acknowledgements.  Such acknowledgements should be placed at the end of the section, in an unnumbered section that does not count towards the paper page limit. Typically, this will include thanks to reviewers who gave useful comments, to colleagues who contributed to the ideas, and to funding agencies and corporate sponsors that provided financial support.

\section*{Impact Statement}

InterBand-EMG and SynapMyo aim to support research on
learning hand-motion representations from wearable EMG.
Potential applications include accessible interaction,
assistive interfaces, and demonstration capture for robotics.
EMG and pose recordings may reveal individual movement
patterns and enable unintended behavioral inference.
Responsible collection, release, and reuse therefore require
appropriate participant consent, de-identification, and
measures to prevent unauthorized identification or monitoring.

Uneven movement coverage and heterogeneous pose annotations
may lead to differences in performance across users and
activities. These limitations should inform downstream
evaluation and dataset reuse. Our experiments evaluate
offline tracking with a provided initial pose; they do not
establish the safety or reliability of real-time control.
Deployment in wearable or robotic systems requires further
validation of latency, temporal stability, robustness to
sensor repositioning, and behavior under prediction errors.

\nocite{langley00}

\bibliography{example_paper}
\bibliographystyle{icml2026}

%%%%%%%%%%%%%%%%%%%%%%%%%%%%%%%%%%%%%%%%%%%%%%%%%%%%%%%%%%%%%%%%%%%%%%%%%%%%%%%
%%%%%%%%%%%%%%%%%%%%%%%%%%%%%%%%%%%%%%%%%%%%%%%%%%%%%%%%%%%%%%%%%%%%%%%%%%%%%%%
% APPENDIX
%%%%%%%%%%%%%%%%%%%%%%%%%%%%%%%%%%%%%%%%%%%%%%%%%%%%%%%%%%%%%%%%%%%%%%%%%%%%%%%
%%%%%%%%%%%%%%%%%%%%%%%%%%%%%%%%%%%%%%%%%%%%%%%%%%%%%%%%%%%%%%%%%%%%%%%%%%%%%%%
\newpage
\appendix
\onecolumn

\input{appendix/appendix_dataset_qc}
\input{appendix/appendix_movement_inventory}
\input{appendix/pose_label_quality_addendum}
\input{appendix/appendix_evaluation_protocol}
\input{appendix/appendix_experimental_setup}
\input{appendix/pretrained_scratch}
\input{appendix/label_efficiency}
\input{appendix/band_information}
\input{appendix/appendix_synapmyo_qualitative}

\end{document}

%% file: appendix/appendix_dataset_qc.tex
\section{Dataset composition and quality-control details}
\label{app:synapmyo_qc}

\paragraph{Inventory construction.}
The source inventory contained 82,887 candidate single-hand exports.
File-level screening excluded 274 sequences taht were too short for the
4,000-sample warm-up as well as three corrupted HDF5 files. A separate metadata
selection rule then excluded a further 920 records whose task annotations explicitly
contained the source-specific bare-hand tag, including  those with mixed
glove/bare-hand task annotations. This tag-based rule is distinct from
the use of vision-derived pose annotations. The resulting processed
inventory contained 81,690 recordings. The current benchmark
excludes a further 72 records with missing or non-action task labels as well as 54 records in overlapping groups that mix training and held-out task assignments. This leaves 81,564 recordings. These metadata exclusions do not constitute an estimate of pose-label validity.

\paragraph{Signal preprocessing.}
The EMG signal is processed at a nominal sampling rate of 2\,kHz using a fourth-order
Butterworth high-pass filter with a 20\,Hz cutoff as well as Butterworth
band-stop filters with stop bands of 48--52, 98--102, and 148--152\,Hz to remove power line interference and its harmonics.
The band-stop filters are constructed with the order parameter set to
four. Filtering is causal, with filter states preserved across processing
chunks. No amplitude normalization is applied when generating these
dataset files. The first 4,000 samples are removed from the EMG, pose, and
other aligned fields. Consequently, an input sequence must
contain more than 4,000 samples to produce a non-empty filtered recording.
%No 450\,Hz low-pass filter is used in this preprocessing pipeline. 
% why to say this?

\paragraph{Conversion checks and temporal handling.}
EMG samples are only retained when they contain the expected eight-channel
vector and a timestamp for the sample. Glove messages must contain 20 ergonomic
angle entries. Exporting requires the relevant EMG and pose streams.
During vision conversion, detections are screened using the upstream
\texttt{quality\_valid} flag. EMG samples are sorted by timestamp.
Samples with the same timestamp are kept, and their timestamps are spread between adjacent timestamp groups. Any remaining non-increasing timestamps
are advanced by the nominal sampling interval of 0.5\,ms. Pose trajectories
are linearly interpolated onto the EMG timestamps, with endpoint values
used outside pose coverage. This interpolation does not impose a maximum
gap between available pose frames. These operations should not be
interpreted as the recovery of missing EMG packets or as an estimate of
hardware synchronization accuracy.

\paragraph{Audit coverage.}
The historical file audit covered all 82,887 candidate exports. For
each of the 81,564 records selected by the current metadata, three windows totalling 768 samples were examined and no non-finite sampled EMG, pose or time values, or sampled non-increasing timestamps were reported.
These are sample checks rather than exhaustive frame-level validation.
The historical glove checks preceded the subsequent dmap2 label update.
Warnings about constant sampled pose values were retained rather than
used as automatic exclusion criteria. Therefore, the audit does not
establish the proportion of valid pose observations in the final dataset.

\paragraph{Separate signal-quality assessment.}
A separate EMG quality report evaluated 56 recordings, corresponding to
448 channel-level observations. It reported a median sampling rate of
1,996\,Hz, with a range of 1,979--2,014\,Hz. Under the scoring protocol of that report, 423 of the 448 channel-level observations (94.42\%) achieved a
quality score of at least 70. However, these statistics only describe the
assessed subset and do not estimate corpus-wide EMG usability or
pose-label validity. The report used its own high-pass and notch filter
configuration, which differs from the dataset preprocessing specified
above. Its scoring formula and the full detector thresholds are not
provided in the available report.

% AUTHOR TODO: Add a citable archived version of the internal device-quality
% report if the separate-assessment paragraph is retained in the submission.
% Do not invent thresholds for saturation or packet-loss rejection:
% no corresponding dataset-wide exclusion rule was established by this audit.

%% file: appendix/appendix_movement_inventory.tex
% Add these packages to the MAIN DOCUMENT PREAMBLE:
% \usepackage{booktabs}
% \usepackage{longtable}
%
% Include this file in the single-column appendix using \input{...}.
% If the appendix is still in two-column mode, place \onecolumn
% at the beginning of the appendix, before including this file.
% Requires \usepackage{array} in addition to booktabs and longtable.
% Original movement labels, counts, and held-out markers are preserved.

\section{Annotated movement inventory}
\label{app:synapmyo_movements}

The inventory comprises 68 gesture and movement categories derived
from recording-level acquisition-task annotations. We normalize
annotation aliases, count explicitly listed components of compound
task labels separately, and exclude stop-control markers.
A recording with multiple annotated components contributes to
each corresponding category; therefore, category-wise recording
counts are not additive. User counts indicate distinct participants.
The V/G column reports the percentages of vision-derived and
glove-derived records within each category, respectively.
These proportions refer to all retained records in the category,
rather than recording duration or only the test subset. These annotations identify recording-level tasks
and do not provide frame-level movement segmentation.

We designate eight categories as held-out tasks: OK gesture
(No.~1), crossed fingers (No.~10), finger wave (No.~28),
thumb MCP flexion (No.~29), thumb circles (No.~37),
rock-horns hand sign (No.~44), writing in the air (No.~58),
and hook fist (No.~62). These categories are marked with
$\dagger$ in Table~\ref{tab:synapmyo_movements}.
Any recording whose normalized annotation includes at least
one designated category, either alone or within a compound
task label, is assigned to a held-out-task subset and excluded
from training.

Validation and test contain the same eight held-out task
categories but use disjoint recording groups. Thus, this
evaluation assesses performance on new recordings of categories
excluded from training but represented during validation.
The exclusion rule operates on annotated tasks and does not
establish the absence of incidental occurrences of these
movements in other recordings.

We additionally report category-wise Tracking performance of the
band-query QKV version of InterBand-EMG, using the same
validation-selected checkpoint as the SynapMyo main table.
For each category and test condition, errors are pooled over valid
frames within each user, then summarized as an equal-user mean
and sample standard deviation across users.
Compound-task records contribute their whole-record errors to each
annotated category; these results are not errors on isolated,
temporally segmented movements. Accordingly, non-held-out categories
can also appear in Task or User+Task through compound annotations.
The test partitions provide scored records for 43 of the 68 categories.
The other 25 categories occur only in training and have no test results.
The Records, Users, and V/G columns describe the full benchmark;
the performance columns use only scored test records.

\begingroup
\footnotesize
\setlength{\tabcolsep}{3pt}
\renewcommand{\arraystretch}{1.15}
\setlength{\LTcapwidth}{\textwidth}
\newcommand{\SynScore}[2]{\ensuremath{#1\,{\scriptstyle\pm#2}}}

\begin{longtable}{@{}r>{\raggedright\arraybackslash}p{0.23\textwidth}rrc*{6}{c}@{}}
\caption{SynapMyo movement coverage, annotation-source composition,
and category-wise Tracking performance of InterBand-EMG (QKV).
V/G gives vision/glove percentages by recording count within each
category. Results are mean $\pm$ sample SD across evaluated users;
MAE is in degrees and LD is in millimeters. Lower is better.
$\dagger$ marks categories excluded from training.
A dash indicates no scored test records for that category and condition;
$\ddagger$ indicates one evaluated user, for whom between-user SD
is undefined. Compound annotations can contribute to multiple rows.}
\label{tab:synapmyo_movements}\\
\toprule
& & & & & \multicolumn{2}{c}{User}
& \multicolumn{2}{c}{Task} & \multicolumn{2}{c}{User+Task} \\
\cmidrule(lr){6-7}\cmidrule(lr){8-9}\cmidrule(l){10-11}
No. & Movement category & Records & Users & V/G (\%)
& MAE & LD & MAE & LD & MAE & LD \\
\midrule
\endfirsthead

\multicolumn{11}{c}{Table~\thetable\ (continued)} \\
\toprule
& & & & & \multicolumn{2}{c}{User}
& \multicolumn{2}{c}{Task} & \multicolumn{2}{c}{User+Task} \\
\cmidrule(lr){6-7}\cmidrule(lr){8-9}\cmidrule(l){10-11}
No. & Movement category & Records & Users & V/G (\%)
& MAE & LD & MAE & LD & MAE & LD \\
\midrule
\endhead

\midrule
\multicolumn{11}{r}{Continued on next page} \\
\endfoot

\bottomrule
\endlastfoot

1 & OK gesture$^{\dagger}$ & 2,111 & 348 & 36.1/63.9 & -- & -- & \SynScore{5.46}{1.70} & \SynScore{8.08}{2.61} & \SynScore{5.96}{2.08} & \SynScore{8.55}{2.36} \\
2 & Counting from one to ten & 2,629 & 347 & 40.5/59.5 & \SynScore{6.36}{1.28} & \SynScore{7.75}{1.95} & \SynScore{5.31}{1.42} & \SynScore{7.14}{2.24} & -- & -- \\
3 & Three-finger gesture (thumb, index, middle) & 1,997 & 339 & 33.8/66.2 & \SynScore{4.83}{1.40} & \SynScore{7.00}{2.43} & $10.39^{\ddagger}$ & $15.20^{\ddagger}$ & -- & -- \\
4 & Middle-finger opposition & 60 & 2 & 0.0/100.0 & -- & -- & -- & -- & -- & -- \\
5 & Middle-finger flexion and extension & 2,140 & 341 & 40.0/60.0 & \SynScore{4.22}{1.62} & \SynScore{6.61}{3.05} & $3.70^{\ddagger}$ & $5.00^{\ddagger}$ & -- & -- \\
6 & Isolated middle-finger extension & 312 & 6 & 85.9/14.1 & -- & -- & -- & -- & -- & -- \\
7 & Bringing fingers together, then spreading them & 268 & 6 & 87.3/12.7 & -- & -- & -- & -- & -- & -- \\
8 & Placing the hand flat with fingers together & 26 & 3 & 0.0/100.0 & -- & -- & -- & -- & -- & -- \\
9 & Spreading and closing the fingers & 988 & 150 & 77.7/22.3 & \SynScore{4.34}{0.56} & \SynScore{6.16}{1.16} & -- & -- & -- & -- \\
10 & Crossed fingers$^{\dagger}$ & 1,174 & 150 & 77.5/22.5 & -- & -- & \SynScore{4.67}{1.27} & \SynScore{7.23}{1.60} & \SynScore{5.11}{1.24} & \SynScore{7.57}{1.42} \\
11 & Fully open hand & 2,266 & 348 & 35.5/64.5 & \SynScore{4.15}{1.60} & \SynScore{6.20}{2.53} & \SynScore{6.89}{3.91} & \SynScore{9.05}{4.51} & -- & -- \\
12 & Full fist at three force levels & 2,518 & 349 & 45.8/54.2 & \SynScore{5.20}{1.45} & \SynScore{6.28}{2.06} & \SynScore{6.38}{2.82} & \SynScore{7.49}{3.56} & \SynScore{6.91}{1.42} & \SynScore{7.81}{1.49} \\
13 & Writing & 16 & 4 & 0.0/100.0 & -- & -- & -- & -- & -- & -- \\
14 & V sign & 2,063 & 347 & 35.3/64.7 & \SynScore{5.58}{1.86} & \SynScore{7.44}{2.58} & $6.54^{\ddagger}$ & $8.41^{\ddagger}$ & -- & -- \\
15 & Beckoning with a finger & 1,134 & 149 & 77.2/22.8 & \SynScore{6.35}{1.19} & \SynScore{7.68}{1.27} & -- & -- & -- & -- \\
16 & Unconstrained random movements of both hands & 1,968 & 145 & 77.5/22.5 & \SynScore{6.35}{1.17} & \SynScore{8.43}{2.06} & -- & -- & -- & -- \\
17 & Separating four fingers into two pairs & 308 & 6 & 88.3/11.7 & -- & -- & -- & -- & -- & -- \\
18 & Separating the little finger from the other three fingers & 314 & 6 & 85.4/14.6 & -- & -- & -- & -- & -- & -- \\
19 & Little-finger abduction & 2,098 & 344 & 39.4/60.6 & \SynScore{1.75}{0.76} & \SynScore{2.86}{0.93} & \SynScore{5.34}{2.10} & \SynScore{8.41}{2.97} & -- & -- \\
20 & Little-finger opposition & 56 & 2 & 0.0/100.0 & -- & -- & -- & -- & -- & -- \\
21 & Little-finger flexion and extension & 2,164 & 340 & 41.7/58.3 & \SynScore{4.30}{1.35} & \SynScore{6.03}{1.80} & $3.77^{\ddagger}$ & $5.16^{\ddagger}$ & -- & -- \\
22 & Little-finger flexion, extension, and abduction & 1,000 & 150 & 76.8/23.2 & \SynScore{4.77}{0.72} & \SynScore{6.88}{1.11} & -- & -- & -- & -- \\
23 & Rapid sequential finger flexion and extension & 1,052 & 149 & 75.9/24.1 & \SynScore{4.03}{0.62} & \SynScore{6.03}{1.33} & -- & -- & -- & -- \\
24 & I-love-you hand sign & 1,240 & 150 & 77.3/22.7 & \SynScore{4.01}{0.58} & \SynScore{5.69}{0.63} & -- & -- & -- & -- \\
25 & Continuous transitions between gestures & 870 & 148 & 77.7/22.3 & \SynScore{4.85}{0.65} & \SynScore{7.01}{1.11} & -- & -- & -- & -- \\
26 & Finger separation & 10 & 2 & 100.0/0.0 & -- & -- & -- & -- & -- & -- \\
27 & Finger wiggling & 1,096 & 150 & 77.6/22.4 & \SynScore{5.29}{0.78} & \SynScore{7.41}{1.29} & -- & -- & -- & -- \\
28 & Finger wave$^{\dagger}$ & 2,577 & 349 & 42.8/57.2 & -- & -- & \SynScore{6.56}{1.79} & \SynScore{9.03}{2.79} & \SynScore{6.78}{1.45} & \SynScore{9.36}{2.22} \\
29 & Thumb MCP flexion$^{\dagger}$ & 2,280 & 345 & 41.1/58.9 & -- & -- & \SynScore{3.35}{1.15} & \SynScore{5.21}{1.77} & \SynScore{3.76}{1.31} & \SynScore{5.79}{2.13} \\
30 & Thumb--middle-finger pinch & 342 & 6 & 87.7/12.3 & -- & -- & -- & -- & -- & -- \\
31 & Thumb-to-finger pinch & 10 & 2 & 100.0/0.0 & -- & -- & -- & -- & -- & -- \\
32 & Four-direction thumb swipes & 950 & 149 & 76.8/23.2 & \SynScore{2.90}{1.10} & \SynScore{4.10}{2.21} & -- & -- & -- & -- \\
33 & Thumb opposition & 2,240 & 345 & 41.4/58.6 & \SynScore{4.54}{0.99} & \SynScore{6.51}{1.59} & \SynScore{2.85}{1.29} & \SynScore{4.26}{1.27} & -- & -- \\
34 & Thumb--little-finger pinch & 326 & 6 & 89.0/11.0 & -- & -- & -- & -- & -- & -- \\
35 & Sweeping the thumb above the four fingers & 918 & 149 & 77.1/22.9 & \SynScore{3.95}{0.66} & \SynScore{5.46}{1.84} & -- & -- & -- & -- \\
36 & Thumb--ring-finger pinch & 336 & 6 & 87.5/12.5 & -- & -- & -- & -- & -- & -- \\
37 & Thumb circles$^{\dagger}$ & 926 & 150 & 76.7/23.3 & -- & -- & \SynScore{3.92}{1.32} & \SynScore{5.81}{1.81} & \SynScore{3.68}{0.87} & \SynScore{5.56}{2.01} \\
38 & Thumb--index-finger pinch & 362 & 7 & 87.3/12.7 & -- & -- & -- & -- & -- & -- \\
39 & Wringing a towel & 10 & 3 & 0.0/100.0 & -- & -- & -- & -- & -- & -- \\
40 & Twisting a bottle cap & 16 & 3 & 0.0/100.0 & -- & -- & -- & -- & -- & -- \\
41 & Pointing & 1,993 & 344 & 34.4/65.6 & \SynScore{5.40}{1.20} & \SynScore{6.98}{1.77} & $4.21^{\ddagger}$ & $4.95^{\ddagger}$ & -- & -- \\
42 & Pointing in different directions & 1,086 & 149 & 77.0/23.0 & \SynScore{5.87}{0.87} & \SynScore{6.98}{1.00} & -- & -- & -- & -- \\
43 & Extending individual fingers from a fist & 968 & 149 & 76.9/23.1 & \SynScore{5.98}{1.51} & \SynScore{6.71}{1.93} & -- & -- & -- & -- \\
44 & Rock-horns hand sign$^{\dagger}$ & 2,021 & 345 & 35.4/64.6 & -- & -- & \SynScore{5.39}{1.36} & \SynScore{7.15}{1.83} & \SynScore{5.78}{1.64} & \SynScore{7.83}{2.13} \\
45 & Number-one gesture & 2,208 & 347 & 37.3/62.7 & \SynScore{5.25}{1.33} & \SynScore{6.76}{2.48} & \SynScore{8.26}{4.38} & \SynScore{10.19}{5.72} & -- & -- \\
46 & Number-two gesture & 2,176 & 346 & 36.7/63.3 & \SynScore{4.84}{1.45} & \SynScore{6.39}{2.01} & -- & -- & -- & -- \\
47 & Number-three gesture & 2,158 & 349 & 36.2/63.8 & \SynScore{4.85}{1.28} & \SynScore{6.74}{1.75} & -- & -- & -- & -- \\
48 & Number-four gesture & 2,082 & 347 & 36.5/63.5 & \SynScore{3.81}{1.16} & \SynScore{5.96}{1.89} & -- & -- & -- & -- \\
49 & Number-five gesture & 2,084 & 345 & 35.9/64.1 & \SynScore{4.29}{1.73} & \SynScore{6.32}{2.55} & -- & -- & -- & -- \\
50 & Ring-finger opposition & 60 & 2 & 0.0/100.0 & -- & -- & -- & -- & -- & -- \\
51 & Ring-finger flexion and extension & 2,122 & 338 & 40.4/59.6 & \SynScore{3.97}{1.64} & \SynScore{6.18}{2.90} & $3.70^{\ddagger}$ & $5.00^{\ddagger}$ & -- & -- \\
52 & Touching the palm with the ring finger & 1,024 & 150 & 77.5/22.5 & \SynScore{4.08}{0.54} & \SynScore{6.18}{1.35} & -- & -- & -- & -- \\
53 & Isolated middle-finger lift from a tabletop & 282 & 6 & 84.4/15.6 & -- & -- & -- & -- & -- & -- \\
54 & Isolated little-finger lift from a tabletop & 282 & 6 & 85.1/14.9 & -- & -- & -- & -- & -- & -- \\
55 & Sequential finger lifts from a tabletop (excluding ring finger) & 298 & 6 & 87.9/12.1 & -- & -- & -- & -- & -- & -- \\
56 & Isolated index-finger lift from a tabletop & 296 & 6 & 85.1/14.9 & -- & -- & -- & -- & -- & -- \\
57 & Thumbs-up gesture & 2,096 & 344 & 36.1/63.9 & \SynScore{5.21}{1.77} & \SynScore{6.44}{2.06} & -- & -- & -- & -- \\
58 & Writing in the air$^{\dagger}$ & 898 & 149 & 76.8/23.2 & -- & -- & \SynScore{4.34}{0.87} & \SynScore{5.02}{1.17} & \SynScore{4.81}{0.90} & \SynScore{5.69}{1.56} \\
59 & Playing an imaginary instrument in the air & 854 & 150 & 77.0/23.0 & \SynScore{4.29}{0.82} & \SynScore{7.24}{2.04} & -- & -- & -- & -- \\
60 & Gesticulating in the air & 856 & 150 & 77.1/22.9 & \SynScore{4.92}{0.84} & \SynScore{5.92}{1.09} & -- & -- & -- & -- \\
61 & Rhythmic hand opening and closing & 2,463 & 349 & 42.6/57.4 & \SynScore{3.60}{1.23} & \SynScore{4.74}{1.59} & \SynScore{4.94}{1.66} & \SynScore{6.53}{2.44} & \SynScore{3.09}{1.10} & \SynScore{3.80}{1.41} \\
62 & Hook fist$^{\dagger}$ & 2,524 & 349 & 44.6/55.4 & -- & -- & \SynScore{6.90}{2.21} & \SynScore{8.18}{2.53} & \SynScore{7.80}{2.09} & \SynScore{9.12}{3.06} \\
63 & Keyboard typing & 14 & 3 & 0.0/100.0 & -- & -- & -- & -- & -- & -- \\
64 & Random movements & 6,059 & 351 & 36.8/63.2 & \SynScore{8.34}{1.43} & \SynScore{11.37}{3.05} & \SynScore{6.85}{2.54} & \SynScore{9.65}{3.75} & -- & -- \\
65 & Separating the index finger from the other three fingers & 312 & 6 & 87.8/12.2 & -- & -- & -- & -- & -- & -- \\
66 & Index-finger opposition & 62 & 3 & 0.0/100.0 & -- & -- & -- & -- & -- & -- \\
67 & Index-finger flexion and extension & 2,136 & 345 & 38.8/61.2 & \SynScore{4.08}{1.56} & \SynScore{5.97}{2.54} & $3.58^{\ddagger}$ & $4.65^{\ddagger}$ & -- & -- \\
68 & Index-finger circles in the air & 266 & 6 & 86.5/13.5 & -- & -- & -- & -- & -- & -- \\

\end{longtable}
\endgroup

%% file: appendix/pose_label_quality_addendum.tex
% Included by appendix_dataset_qc.tex after temporal handling.
% Requires no extra package. Do not include a second time in the main text.
\section{Pose-label screening and numerical checks}
\label{app:synapmyo_label_quality}

\paragraph{Source-observation screening.}
Before temporal alignment, vision observations for a hand are
rejected when the hand is not detected or its \texttt{quality \_valid} flag is false. Glove messages are rejected when their hand-side field is neither \texttt{left} nor \texttt{right}, or their ergonomics list does not contain exactly
20 entries.

We audit the complete source spans associated with the 81,564 benchmark recordings, considering the hand sides retained in the dataset before interpolation and EMG cropping.
The source-observation acceptance rate is the fraction of audited observation opportunities passing the corresponding source-specific checks. For vision, the denominator includes cached frame opportunities for each retained hand, including
missed detections. For glove data, it includes native messages
associated with retained hand sides.
The accepted and total counts are 59,821,253/79,784,241 for vision and 239,090,005/239,090,005 for glove, corresponding to acceptance rates of 74.98\% and 100.00\%, respectively.
All source episodes were matched successfully, with no reading failures. These rates describe source-specific screening outcomes; they do not measure joint-angle accuracy or the fraction of recording time supported by reliable pose observations.

\paragraph{Labels used for supervised learning.}
Accepted source observations are converted to the common joint-angle representation and linearly interpolated onto EMG timestamps. Rejecting a source observation does not necessarly remove the corresponding EMG samples from supervised learning: the corresponding interval may receive labels interpolated from neighboring accepted observations. Endpoint values are used outside the available pose interval, as described in Appendix~\ref{app:synapmyo_qc}. 

The supervised loader reads the resulting dense joint-angle
sequences. It checks for finite values and constructs windows
from consecutive frames whose 20-dimensional angle vectors are not all numerically close to zero. Non-finite values trigger an error rather than silent frame removal. The loader does not read the source quality flags or source-pose timestamps, and it applies no maximum interpolation-gap rule. Self-supervised pretraining reads EMG without pose labels.

\paragraph{Coverage of dense-label numerical checks.}
The completed supervised training and validation index covers
69,557 recordings and 7,493,979,977 stored pose rows.
Finite-value checks completed successfully, and no rows
failed the all-near-zero check. Of these recordings, 434
yield no complete supervised windows under the 11,790-sample
window and endpoint rules.

These checks establish numerical consistency for the indexed
training and validation data. They do not include the 12,007
test recordings, establish independent pose accuracy, or imply
that every indexed row contributes to a training loss.
Corpus-wide temporal support under an interpolation-gap
constraint and independent pose accuracy remain unmeasured.

% AUTHOR NOTE: The 50/100/200 ms interpolation-gap settings in the separate
% pilot audit are sensitivity analyses only. They were not used to generate
% the released labels, select training windows, or mask training losses.
% Corpus-wide gap-constrained temporal support and independent pose accuracy
% remain unmeasured; do not substitute the source-observation pass rate for either.

%% file: appendix/appendix_evaluation_protocol.tex
\section{Evaluation protocol}
\label{app:synapmyo_evaluation}

\paragraph{Task and window construction.}
We evaluate initial-pose-conditioned tracking of 20 joint
angles. Gesture categories define recording annotations and
evaluation splits; they are not prediction targets.

At the nominal EMG sampling rate of $2\,\mathrm{kHz}$,
each input window contains 11,790 samples: 1,790 context
samples followed by 10,000 scored samples. Using zero-based
window indices, the supplied initial reference pose is at
index 1,790, and scoring covers indices 1,790--11,789.
No subsequent reference poses are supplied to the model.
Model state is reset for each window.

Evaluation uses a stride of 11,790 samples without random
offsets. Scored intervals are therefore separated by the
next window's context interval. Only complete windows within
valid continuous runs are retained. The loader excludes the
last two samples of each run from the eligible interval;
a run of length $L$ therefore contributes
\begin{equation}
N(L)=\max\!\left(
0,\left\lfloor\frac{L-2}{11{,}790}\right\rfloor
\right)
\end{equation}
windows. Incomplete intervals are not evaluated.

\paragraph{Temporal resampling and angular error.}
The decoder produces 250 prediction steps per window at a
nominal rate of $50\,\mathrm{Hz}$. For angular evaluation,
these outputs are resized to 10,000 samples using linear
interpolation with \texttt{align\_corners=false}.
This is index-based resampling; the dense grid does not
represent independent pose predictions at $2\,\mathrm{kHz}$.

Let $\mathcal{U}_g$ denote the users evaluated in
generalization condition $g$, and let $\mathcal{E}_{g,u}$
contain the valid evaluated window--frame pairs belonging
to user $u$. For predicted and reference joint angles
$\hat{\theta}_{w,t,j}$ and $\theta_{w,t,j}$ in radians,
the angular error for user $u$ is
\begin{equation}
\mathrm{MAE}_{g,u}
=
\frac{180}{20\pi\lvert\mathcal{E}_{g,u}\rvert}
\sum_{(w,t)\in\mathcal{E}_{g,u}}
\sum_{j=1}^{20}
\left|
\hat{\theta}_{w,t,j}-\theta_{w,t,j}
\right|.
\end{equation}
All 20 joint-angle dimensions, including thumb MCP flexion,
are included. The factor $180/\pi$ converts radians to degrees.

Errors are pooled across all evaluated windows and recordings
within each user. We then report the equally weighted mean
and sample standard deviation across users. With
$N_g=|\mathcal{U}_g|$, these are
\begin{equation}
\begin{aligned}
\overline{\mathrm{MAE}}_g
&=
\frac{1}{N_g}
\sum_{u\in\mathcal{U}_g}\mathrm{MAE}_{g,u},\\
s_g
&=
\sqrt{
\frac{1}{N_g-1}
\sum_{u\in\mathcal{U}_g}
\left(
\mathrm{MAE}_{g,u}-\overline{\mathrm{MAE}}_g
\right)^2
}.
\end{aligned}
\end{equation}
Each user therefore contributes equally to the condition-level
score, and the standard deviation describes variation across users.

\paragraph{Landmark distance.}
Predicted and reference angles are converted to landmark
positions using the same fixed UmeTrack template,
\texttt{bones\_model\_noe1.json}, in millimeter-scale coordinates.
Two zero wrist angles are appended for forward kinematics,
and the global wrist transformation is the identity.
No participant-specific scaling or sequence-specific rigid
alignment is applied. The metric therefore measures
angle-induced positional differences under a common geometry.

Following the emg2pose landmark selection, we exclude
zero-based indices 8, 11, 14, and 17, corresponding to the
four non-thumb proximal landmarks. The remaining 17 landmarks,
including the wrist and palm center, are
listed in Table~\ref{tab:synapmyo_eval_landmarks}.
Landmark evaluation selects every fortieth sample from the
interpolated prediction and aligned reference trajectories,
starting at the first scored sample. This yields 250 evaluated
frames per window at a nominal rate of $50,\mathrm{Hz}$.
Let $\mathcal{E}^{50}g$ denote these window--frame pairs,
$\mathcal{K}$ the selected landmark set, and
$\hat{p}{w,t,k}$ and $p_{w,t,k}$ the corresponding positions.

\begin{table}[t]
    \centering
    \caption{Landmarks included in the distance metric. Indices
    follow the zero-based ordering of the UmeTrack implementation.}
    \label{tab:synapmyo_eval_landmarks}
    \small
    \begin{tabular}{@{}rl@{}}
        \toprule
        Index & Landmark \\
        \midrule
        0  & Thumb fingertip \\
        1  & Index fingertip \\
        2  & Middle fingertip \\
        3  & Ring fingertip \\
        4  & Little fingertip \\
        5  & Wrist joint \\
        6  & Thumb intermediate landmark \\
        7  & Thumb distal landmark \\
        9  & Index intermediate landmark \\
        10 & Index distal landmark \\
        12 & Middle intermediate landmark \\
        13 & Middle distal landmark \\
        15 & Ring intermediate landmark \\
        16 & Ring distal landmark \\
        18 & Little intermediate landmark \\
        19 & Little distal landmark \\
        20 & Palm center \\
        \bottomrule
    \end{tabular}
\end{table}

The mean landmark distance is
\begin{equation}
    \mathrm{LD}_g =
    \frac{1}{17\lvert\mathcal{E}^{50}_g\rvert}
    \sum_{(w,t)\in\mathcal{E}^{50}_g}
    \sum_{k\in\mathcal{K}}
    \left\lVert\hat{p}_{w,t,k}-p_{w,t,k}\right\rVert_2.
\end{equation}
Distances are reported in millimeters. For fingertip distance, $\mathcal{K}={0,1,2,3,4}$ and
the denominator uses 5 instead of 17.

\paragraph{Validity checks and aggregation.}
Non-finite reference angles trigger an error.
Frames whose 20 angles are all within $10^{-8}$ radians
of zero are excluded from valid runs; individual zero angles
remain eligible. These are numerical eligibility checks,
not measures of pose accuracy.
Source quality flags are applied during label construction,
as described in
Appendix~\ref{app:synapmyo_label_quality} and are not reapplied during evaluation.
No maximum interpolation-gap threshold is imposed.

We report held-out-user, held-out-task, and combined-condition
scores separately. Within each condition, valid-frame errors
are pooled across all evaluated windows and recordings of
each user to obtain one score per user. For a metric
$e_{g,u}\in\{\mathrm{MAE}_{g,u},\mathrm{LD}_{g,u},
\mathrm{FD}_{g,u}\}$ and $N_g=|\mathcal{U}_g|$, we report
\begin{equation}
\bar e_g=\frac{1}{N_g}\sum_{u\in\mathcal{U}_g}e_{g,u},
\qquad
s_g=
\sqrt{\frac{1}{N_g-1}
\sum_{u\in\mathcal{U}_g}(e_{g,u}-\bar e_g)^2}.
\end{equation}
Thus, users receive equal weight, and $s_g$ measures variation
across users. Angular and spatial errors use their respective
2\,kHz and 50\,Hz evaluation grids. Windows duplicated solely
for distributed-batch padding are counted once.
For SynapMyo, checkpoint selection minimizes the arithmetic
mean of the three frame-averaged validation-condition angular
MAEs. 

%% file: appendix/appendix_experimental_setup.tex
% Requires booktabs. No additional packages are needed beyond the usual
% ICML template packages. Insert after the existing \appendix command.
% Do not add a second \appendix command for this file.
% TODOs are intentionally visible; they are not verified run parameters.

\section{Experimental Settings}
\label{app:experimental_setup}

\subsection{Data partitions and evaluation}
\label{app:experiment_protocol}

We use the official emg2pose partitions~\citep{salter2024emg2pose}
and the SynapMyo partitions defined in the benchmark section.
Models are trained separately on the two datasets. Tokenizer
training, encoder pretraining, and pose fine-tuning use the
corresponding training partition; the first two stages do not
use pose labels. Statistics used to normalize reconstruction
errors are estimated from training EMG. The Task condition
corresponds to the Stage condition in the emg2pose literature.

For our evaluations, each Tracking window contains five seconds of scored
motion, preceded by 1,790 EMG samples of context at 2\,kHz.
The initial ground-truth pose is supplied at the start of the
scored interval. The recurrent decoder operates at 50\,Hz,
and the predicted joint angles are interpolated to the label
grid for angular evaluation. Angular MAE uses all 20 joint
angles and is reported in degrees. Landmark distance uses
the 17 evaluation landmarks at 50\,Hz and is reported in
millimeters. For InterBand-EMG and all SynapMyo results, errors are first
averaged over each user's valid scored frames, then summarized
by the equally weighted mean and sample standard deviation
across users. The context interval is excluded from scoring.
The remaining emg2pose entries retain the aggregation stated
in Table~\ref{tab:emg2pose_main} and
Appendix~\ref{app:published_baselines}.

\subsection{InterBand-EMG architecture}
\label{app:interband_configuration}

Tokenizer training and masked pretraining use 1,800-sample
EMG windows, divided into nine non-overlapping patches of
200 samples per channel. Each band encoder has 12 Transformer
blocks, an embedding dimension of 200, 10 attention heads,
and an MLP expansion ratio of four. The low-, mid-, and
high-frequency codebooks each contain 1,024 entries, with
embedding dimensions of 16, 16, and 32, respectively.
The quantizers use an exponential-moving-average decay of
0.99 and a commitment coefficient of one. Codebooks are
initialized with 10 iterations of k-means using at most
32,768 samples per codebook.

Post-quantization band features are projected to dimension
200. Cross-band attention uses 10 heads, followed by a
shared three-block Transformer decoder with dimension 200
and 10 heads. The tokenizer reconstructs standardized FFT
amplitudes and phases with a phase-loss coefficient of one.
The masked EMG encoder has the same width, depth, and number
of heads as each band encoder, with three separate
1,024-class prediction heads. During pose fine-tuning, the
encoder is updated jointly with a Transformer-based recurrent
decoder of width 256, depth four, and eight attention heads.
Dropout is set to zero. The pose objective is the joint-angle
MAE plus 0.01 times the fingertip-distance loss.

\subsection{Structured masking}
\label{app:structured_masking_settings}

We adapt four mask primitives from the N-MET strategy of
EEG-JEPA~\citep{li2026eeg},
omitting its two EEG-specific topological-pair primitives.
Masks are constructed on the channel--patch grid. Each
candidate region is sampled from temporal, channel, local,
or full-channel masking with probabilities 0.425, 0.2125,
0.2125, and 0.15, respectively. Temporal regions span
25--60\% of temporal patches and 45--90\% of channel rows;
channel regions span 60--100\% of temporal patches and
10--35\% of channel rows. Local regions use 6--18\% of
the grid area, with a channel-to-time aspect parameter
between 0.15 and 0.80. Full-channel regions select
5--16\% of channels across the entire window.
These ranges describe candidate regions before integer
rounding and adjustment to the masking budget.

Candidate regions are combined and adjusted to mask exactly
half of the tokens: 72 of 144 for emg2pose and 36 of 72
for SynapMyo. A mask and its complement form two views
processed by the same encoder, and all three prediction
heads use the same mask. Cross-entropy is averaged across
the three codebooks within each view and then summed
across the two views. Validation uses random masks with
a base seed of 1042. Contiguous channel regions refer to
input channel order; no additional electrode-pair topology
is imposed.

\subsection{Optimization and checkpoint selection}
\label{app:optimization_settings}

Table~\ref{tab:emg2pose_optimization} gives the InterBand-EMG
configuration for emg2pose. All three stages use AdamW and
step-wise linear warm-up followed by cosine learning-rate
decay. Fine-tuning uses encoder layer-wise learning-rate
decay with a factor of 0.65. Biases, one-dimensional
parameters, and designated embedding parameters are
excluded from weight decay during masked pretraining and
fine-tuning. No gradient accumulation is used.

\begin{table}[t]
    \centering
    \caption{InterBand-EMG optimization on emg2pose.
    Stages 1, 2, and 3 denote tokenizer training, masked
    pretraining, and pose fine-tuning. Batch sizes are
    global across 32 GPUs.}
    \label{tab:emg2pose_optimization}
    \small
    \setlength{\tabcolsep}{3pt}
    \renewcommand{\arraystretch}{1.08}
    \begin{tabular}{@{}lrrr@{}}
        \toprule
        Setting & Stage 1 & Stage 2 & Stage 3 \\
        \midrule
        Epochs & 100 & 50 & 50 \\
        Batch size & 6,144 & 3,072 & 128 \\
        Peak LR & $9.798\!\times\!10^{-4}$ & $1.225\!\times\!10^{-3}$ & $5\!\times\!10^{-4}$ \\
        Minimum LR & $10^{-5}$ & $10^{-5}$ & $10^{-6}$ \\
        Weight decay & $10^{-4}$ & 0.05 & 0.05 \\
        Adam $\beta_1$ & 0.9 & 0.9 & 0.9 \\
        Adam $\beta_2$ & 0.99 & 0.98 & 0.999 \\
        Warm-up epochs & 10 & 5 & 5 \\
        Gradient clipping & None & 3 & 1 \\
        Seed & 20260809 & 20260809 & 42 \\
        Precision & BF16 & BF16 & BF16 \\
        \bottomrule
    \end{tabular}
\end{table}

The emg2pose experiments use 32 NVIDIA RTX PRO 6000 GPUs
with BF16 mixed precision and TF32 disabled. FFT target
construction and phase reconstruction errors are evaluated
in FP32. AdamW uses $\epsilon=10^{-8}$; the Stage~3
betas are the optimizer defaults. Training uses temporal
window jitter and cyclic channel rotation, sampled uniformly
from $\{-2,-1,0,1,2\}$ in the first two stages and
$\{-1,0,1\}$ during pose fine-tuning. Validation and test
inputs are not augmented.
The cross-band attention module and phase reconstruction
heads use fixed initialization seeds of 20260916 and
20261012, respectively.

For emg2pose, the final tokenizer checkpoint initializes
Stage~2, and the final pretrained encoder checkpoint
initializes Stage~3. The fine-tuned model is selected by
minimum validation pose loss, using the same weighted
angle and fingertip objective as in training.

For SynapMyo, we use the amplitude-and-phase reconstruction
objective and structured masking specified above.
\texttt{TODO:} Insert the revised eight-channel run's
stage-wise epoch budgets, global batch sizes, learning rates,
weight decay, warm-up and minimum learning rates, clipping,
seeds, precision, hardware, and checkpoint-selection and
handoff rules. These settings must come from the revised
run configuration.
% Do not fill this TODO from the older amplitude-only/random-mask
% public8_acf_fp32_pro6000_6x8_20261010_lb01 run.

\subsection{Supervised baselines on SynapMyo}
\label{app:synapmyo_baselines}

The emg2pose and vemg2pose Tracking architectures are
available in the official emg2pose implementation
\citep{salter2024emg2pose}. Both use a time-depth separable
convolutional encoder with 64-dimensional output features,
concatenated with the preceding 20-dimensional pose.
The emg2pose decoder is an MLP with two hidden layers of
width 512 that predicts absolute joint angles. The
vemg2pose decoder is a two-layer LSTM with hidden width
512 that predicts angular increments, which are accumulated
to obtain the next pose. Both receive the initial pose.
For eight-channel input, the first input convolution is
adapted from 16 to eight channels, while the internal
feature dimensions and 20-dimensional pose output are
retained. The baselines are initialized from scratch.

The recorded SynapMyo vemg2pose recipe uses Adam with a
constant learning rate of $1.5\times10^{-3}$, a global
batch size of 768, gradient-norm clipping at one, and
20 training epochs in FP32 on 48 GPUs. It uses seed 42,
the same angle-plus-fingertip training loss, and the
SynapMyo Tracking windows. Its checkpoint is selected by
the mean angular MAE across the three validation conditions,
with each condition weighted equally.

The eight-channel emg2pose baseline is trained from scratch
for 20 epochs with seed 42, using Adam at a constant learning
rate of $1.5\times10^{-3}$ and gradient-norm clipping at one.
Training uses 32 NVIDIA RTX PRO 6000 GPUs in FP32 with
TF32 disabled, a batch size of 24 per GPU, and a global
batch size of 768, without gradient accumulation.
Each training window contains 11,790 EMG samples:
1,790 context samples followed by a 10,000-sample
prediction interval. The model receives the reference
pose at the start of that interval and is trained with
the same angle-plus-fingertip objective.
Checkpoint selection minimizes the arithmetic mean of
the three frame-averaged validation-condition angular
MAEs.

\subsection{Published results on emg2pose}
\label{app:published_baselines}

The emg2pose comparison includes published Tracking results
for vemg2pose~\citep{salter2024emg2pose},
EMG-GPT~\citep{magni2026emg}, and
NHN~\citep{wang2026neuromotor}. We mark these entries with
$\dagger$ and retain the values reported by their authors.
The vemg2pose results are taken from the original emg2pose
paper. EMG-GPT results are taken from Appendix~E.2,
Table~4 of its paper, and NHN results from Table~1 of
the specified arXiv v1 paper. The selected rows all
evaluate Tracking with a provided initial pose.

EMG-GPT uses six seconds of input, including one second
of preceding context, for a five-second scored interval.
Its Tracking protocol uses 50\,Hz pose targets,
zero-order-hold feature alignment, and a 40\,ms
token-to-pose offset. Table~4 aligns its target indices,
validity checks, metrics, and aggregation with a locally
reevaluated vemg2pose baseline, while retaining each
method's native signal-processing frontend. We do not
substitute that locally reevaluated baseline for the
original paper's vemg2pose result.

NHN is a task-supervised model. Its reported evaluation
uses five-second windows and 50\,Hz rollout. It averages
each user's results over five seeds before computing the
mean and standard deviation across users. EMG-GPT also
reports an equally weighted user mean and user-level
standard deviation. These user-level statistics differ
from the frame-weighted aggregation used in our evaluations;
the reported deviations are not uncertainty across seeds
alone. Accordingly, the published entries retain their
source protocols rather than being described as our
reimplementations under a common evaluation pipeline.

% Suggested main-result table note:
% $\dagger$ Published Tracking results, retained under their source
% protocols; see Appendix~\ref{app:published_baselines}.

%% file: appendix/pretrained_scratch.tex
% Requires \usepackage{graphicx} in the main document.
% Include after \appendix using \input{appendix/pretrained_scratch}.
% Source: output/pretraining_vs_scratch_20261024/comparison_values.json.
% Figure: user means and between-user sample SD, not seed SD.
% The separate 30-epoch comparison is reported in the text as frame means.
\section{Pretraining versus Training from Scratch}
\label{app:pretrained_scratch}

We assess the contribution of pretraining on emg2pose by comparing
InterBand-EMG with the same pose-estimation architecture trained
from scratch. Both conditions use the same EMG encoder and
Transformer-based recurrent decoder, training partition, pose
objective, data augmentation, and downstream optimization recipe.
The pretrained condition initializes the encoder from masked code
prediction, whereas the scratch condition initializes it randomly
and omits tokenizer training and masked pretraining. Both conditions
use downstream seed 42 and identical initial decoder weights, and
jointly optimize the encoder and decoder.

\begin{figure*}[t]
    \centering
    \includegraphics[width=\textwidth]{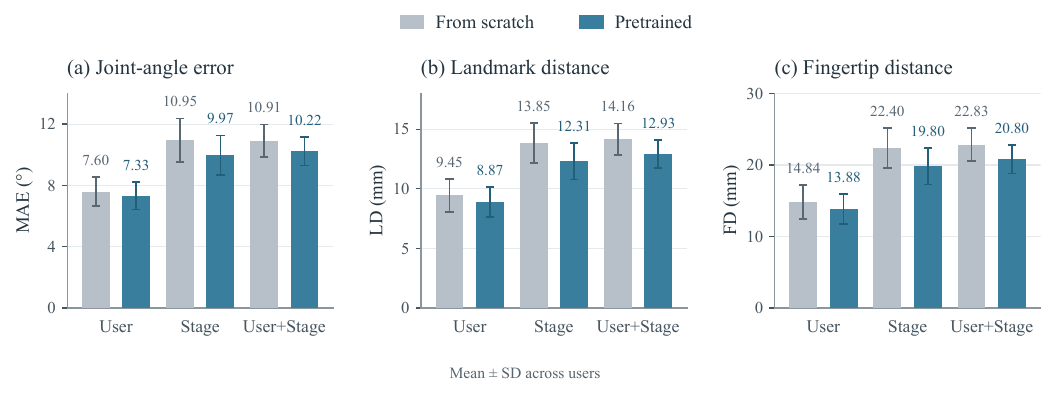}
    \caption{Pretraining versus training from scratch on emg2pose
    Tracking. Both conditions use the same downstream architecture
    and a provided initial pose. Panels show (a) joint-angle MAE,
    (b) distance over 17 landmarks, and (c) distance over five
    fingertips. Bars and numerical labels indicate equal-user
    means; error bars show sample SD across users
    (20/158/20 users for User/Stage/User+Stage).
    Lower values are better.}
    \label{fig:pretrained_scratch}
\end{figure*}

Figure~\ref{fig:pretrained_scratch} reports the available
user-level comparison. For each metric, valid-frame errors are first
averaged within each user; we then report the equally weighted
mean and sample standard deviation across users. The User,
Stage, and User+Stage conditions contain 20, 158, and 20 evaluated
users, respectively. Relative MAE reductions are
3.6\%, 8.9\%, and 6.4\% for User, Stage, and User+Stage,
respectively. The corresponding fingertip-distance reductions
are 6.5\%, 11.6\%, and 8.9\%. These results support the
usefulness of the pretrained representation beyond the choice
of downstream architecture.

%% file: appendix/label_efficiency.tex
% Requires \usepackage{graphicx} in the main document.
% Include after \appendix using \input{appendix/label_efficiency}.
% Verified source: output/label_efficiency_20261026/pretrained_only_results.json.
% Statistics are equal-user means and sample SD across users, not across seeds.
\section{Label Efficiency}
\label{app:label_efficiency}

We examine how the downstream generalization of InterBand-EMG
changes with the amount of pose supervision. All runs initialize
the encoder from the same masked code-prediction checkpoint and
fine-tune the encoder and pose decoder using $10\%$, $20\%$,
$50\%$, or $100\%$ of the training recording groups.
These fractions apply only to supervised fine-tuning; the
tokenizer and encoder pretraining are held fixed.
We evaluate Tracking with a provided initial pose on the same
User, Stage, and User+Stage test partitions at every fraction.

The subsets are nested and sampled deterministically within
the official training partition, stratified by user. They
contain 888, 1,776, 4,441, and 8,882 recording groups,
respectively. Training recordings from the same group are
selected jointly; validation and test data remain unchanged.
The percentages therefore refer to groups, rather than exact
fractions of recording duration or users. All runs use
supervised training seed 42 and the same architecture and loss.
The global batch size and learning rate are 128 and
$5\times10^{-4}$ at $10\%$ and $100\%$, and 256 and
$7.07\times10^{-4}$ at $20\%$ and $50\%$.

\begin{figure}[!h]
    \centering
    \includegraphics[width=0.5\columnwidth]{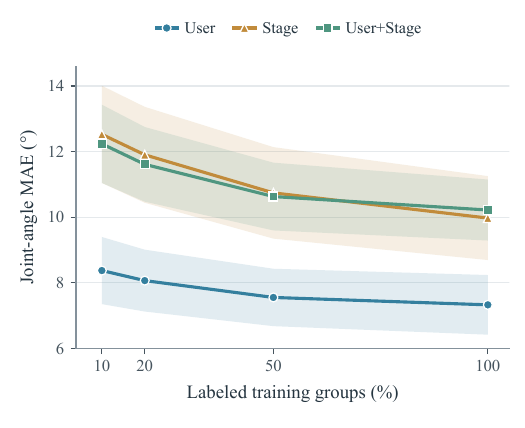}
    \caption{Generalization of InterBand-EMG with limited
    pose labels on emg2pose Tracking. The same pretrained
    encoder initialization is fine-tuned using $10\%$,
    $20\%$, $50\%$, and $100\%$ of the training recording
    groups. Solid lines connect equal-user mean joint-angle
    MAE under the User, Stage, and User+Stage conditions;
    shaded regions show $\pm$ one sample standard deviation
    across users. Lower values are better.}
    \label{fig:label_efficiency}
\end{figure}

Figure~\ref{fig:label_efficiency} shows the equally weighted
mean joint-angle MAE and sample standard deviation across
users, after pooling valid-frame errors within each user.
There are 20, 158, and 20 test users in the User, Stage,
and User+Stage conditions, respectively. MAE decreases
monotonically as more labeled groups are used in all three
conditions. At $50\%$ labels, the model achieves MAEs of
$7.55^\circ$, $10.74^\circ$, and $10.63^\circ$, compared
with $7.33^\circ$, $9.97^\circ$, and $10.22^\circ$ at
$100\%$. Halving the labeled training groups therefore
increases MAE by only $0.23^\circ$, $0.77^\circ$, and
$0.41^\circ$, respectively, under these training configurations.

%% file: appendix/band_information.tex
% Requires \usepackage{booktabs,graphicx} in the main document.
% Include after \appendix using \input{appendix/band_information}.
% Verified against archived user_errors.json and shift_user_errors.json.
% Statistics: equal-user mean and sample SD, not frame SD or seed SD.
\section{Predictive Information across EMG Frequency Bands}
\label{app:band_information}

We examine whether the low-, mid-, and high-frequency components
of EMG provide complementary information for hand-pose estimation.
To separate this analysis from the learned tokenizer, we extract
fixed spectral features and fit linear ridge readouts on emg2pose.
The analysis compares all seven nonempty band combinations,
measures the effect of removing each band, and tests sensitivity
to temporal misalignment. We repeat the analysis with overlapping
and disjoint frequency windows to assess the role of shared
boundary frequencies.

\subsection{Diagnostic protocol}
\label{app:band_information_protocol}

We retain the official validation and test partitions and remove
540 training recordings whose paired-hand recording groups also
occur in a later partition. This conservative group check leaves
16,596 training, 1,950 validation, and 6,167 test recordings; the
diagnostic training subset therefore differs from the main
Tracking benchmark. We select up to 32 uniformly spaced anchor
locations per recording on a 100\,ms grid, using recording length
alone. Each anchor uses the preceding 0.3\,s of EMG. An anchor is
excluded if this context contains nonfinite EMG or pose values,
an all-zero invalid pose, non-increasing timestamps, or a timestamp
gap exceeding $2.5/2000$\,s. The same valid anchors are used for
every band combination. Evaluation covers 165,689
anchors from 6,114 test recordings; the remaining 53 recordings
have no valid anchors. The User, Stage, and User+Stage conditions
contain 20, 158, and 20 users, respectively.

The EMG context is divided into three consecutive 200-sample
patches at 2\,kHz. We subtract each patch's mean, compute a
rectangular-window FFT, and extract four fixed log-power sub-bins
per frequency band and electrode. The nominal bands are
0--150\,Hz, 150--300\,Hz, and 300--450\,Hz. The overlapping
windows use cosine/sine transitions over 125--175\,Hz and
275--325\,Hz, with an upper taper over 420--450\,Hz. Squared
window memberships weight the power spectrum. The disjoint
alternative uses $[0,150)$, $[150,300)$, and $[300,450)$\,Hz,
retaining the same upper taper. Sub-bins divide each band's
support into four equal-width intervals. Each band contributes
$3\times16\times4=192$ features; combinations concatenate the
selected features. Thus, all readouts use the same linear model
family, while their input dimensionality varies with the number
of bands.

For each subset of bands, a linear ridge readout predicts the
current 20 joint angles from the selected spectral features.
The seven inputs are $L$, $M$, $H$, $L+M$, $L+H$, $M+H$,
and $L+M+H$. Each readout receives only the selected band features
and is fitted independently using the same training and validation
procedure.

Feature normalization and readout fitting use training data only,
with equal total weight assigned to each training user. For every
band combination, ridge regularization is selected from
$\{10^{-5},10^{-4},10^{-3},10^{-2},10^{-1},1,10\}$ using the
equally weighted mean of the available validation conditions'
user-averaged MAEs. The official validation metadata provides
User and User+Stage conditions. All fitted readouts are frozen
before test evaluation. At each anchor, we average the absolute
errors over the 20 angles, then pool anchors within each user.
Reported MAEs are the equal-user mean and sample standard
deviation, in degrees.

\subsection{Band combinations and conditional contributions}
\label{app:band_information_contribution}

Table~\ref{tab:band_information_combinations} reports all seven
band combinations. With overlapping windows, combining all three
bands yields the lowest mean error among the seven combinations, achieving
$16.641^\circ$, $19.386^\circ$, and $19.240^\circ$ under User,
Stage, and User+Stage. Relative to the low-frequency band alone,
the corresponding reductions are $0.272^\circ$, $0.101^\circ$,
and $0.112^\circ$. The full-band readout also has the lowest
mean error under disjoint windows, showing that this pattern
does not depend solely on shared boundary frequencies.

To quantify the contribution of band $b$ given the other two,
we compare $M+H$ against $L+M+H$ for the low band, $L+H$ against
$L+M+H$ for the mid band, and $L+M$ against $L+M+H$ for the high
band. We independently refit each reduced readout and compute
\begin{equation}
    \Delta_b = \frac{1}{|\mathcal{U}|}
    \sum_{u\in\mathcal{U}}
    \left[e_u(\mathcal{B}\setminus\{b\})-e_u(\mathcal{B})\right],
    \label{eq:band_information_delta}
\end{equation}
where $\mathcal{B}=\{L,M,H\}$, $\mathcal{U}$ is the set of
users in one evaluation condition, and $e_u$ is that user's MAE.
A positive value indicates lower error when the band is retained.
We estimate marginal 95\% confidence intervals by resampling
paired user differences 2,000 times. Two-sided sign-flip tests
use 10,000 randomizations and assume symmetric user differences
under the null. For the nine band-removal contrasts reported here,
we retain the conservative Holm adjustment from the original
18-contrast analysis within each window definition. Confidence
intervals are not adjusted for multiplicity;
Table~\ref{tab:band_information_removal}
reports the adjusted $p$ values separately.

All mean removal effects are positive, but their statistical
support varies across conditions. For the overlapping-window
readout, retaining each of the three bands improves User MAE
after Holm correction ($p<0.05$). The mid- and high-frequency
effects also meet this criterion under Stage, while none does
under User+Stage. With disjoint windows, the corresponding
effects are significant for low and mid frequencies under User,
all three bands under Stage, and mid frequencies under
User+Stage. These results support complementary predictive
information across the three ranges, with stronger evidence in
some generalization conditions than others.

\input{tables/band_information_combinations}
\input{tables/band_information_removal}

\subsection{Sensitivity to temporal alignment}
\label{app:band_information_alignment}

We additionally perturb the frozen full-band readout by circularly
shifting one band's feature trajectory within each recording.
All electrode, sub-bin, and context-patch features belonging to
that band are shifted together; the remaining bands and pose
targets stay fixed. A joint shift of all bands preserves their
mutual alignment while changing their
alignment with pose. We use 32 shifts per recording, with a
minimum circular displacement of 5\,s. This analysis requires
at least eight regularly spaced valid anchors and therefore
uses 1,790 eligible recordings, covering 20, 146, and 19 users
in the three conditions. Observed and perturbed errors are
computed on exactly the same eligible anchors. Intervals are
the 2.5th--97.5th percentiles across the 32 perturbations,
rather than confidence intervals across users.

\begin{figure*}[t]
    \centering
    \includegraphics[width=\textwidth]{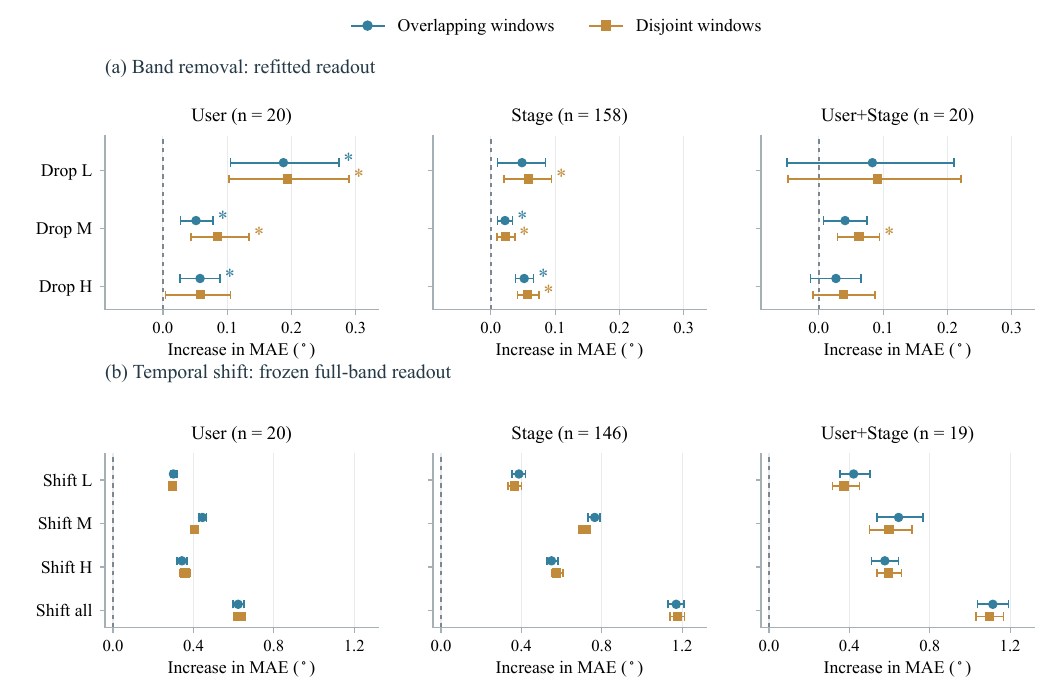}
    \caption{Frequency-band contributions on emg2pose using
    fixed spectral features and linear ridge readouts.
    (a) MAE increase after removing one band and refitting the
    readout. Error bars show marginal paired-user bootstrap
    95\% confidence intervals; asterisks indicate
    Holm-adjusted $p<0.05$, retaining the original
    18-contrast correction within each window definition.
    (b) MAE increase after circularly shifting one band,
    or all bands jointly, with the full-band readout
    frozen. Error bars show the 2.5th--97.5th percentiles over
    32 shifts on matched eligible anchors. They describe
    perturbation variability, not user-level confidence.
    Positive changes indicate lower error when the original
    band information and temporal alignment are retained.
    Overlapping and disjoint windows are shown throughout.}
    \label{fig:band_information}
\end{figure*}

Figure~\ref{fig:band_information} shows that shifting any one
band increases mean prediction error in every condition under
both window definitions. For overlapping windows, joint shifts
increase MAE by $0.623^\circ$, $1.170^\circ$, and $1.114^\circ$.
Positive mean effects also persist with disjoint windows.
Temporal alignment therefore contributes
to the predictive value of these features, although shifting
one band also changes its relationship with the others.

Together, these diagnostics motivate retaining low-, mid-, and
high-frequency information for pose representation learning.
They characterize information accessible to fixed spectral
features and linear readouts.

%% file: tables/band_information_combinations.tex
% Automatically generated from verified archived results.
\begin{table*}[t]
\centering
\footnotesize
\setlength{\tabcolsep}{3pt}
\renewcommand{\arraystretch}{1.08}
\caption{All seven frequency-band combinations on emg2pose using fixed spectral features and independently fitted ridge readouts. Values are joint-angle MAE in degrees, reported as equal-user mean $\pm$ sample SD across users. The User, Stage, and User+Stage conditions contain 20, 158, and 20 users, respectively. Both frequency-window definitions use the same valid samples.}
\label{tab:band_information_combinations}
\begin{tabular*}{\textwidth}{@{\extracolsep{\fill}}l*{6}{c}@{}}
\toprule
& \multicolumn{3}{c}{Overlapping windows} & \multicolumn{3}{c}{Disjoint windows} \\
\cmidrule(lr){2-4}\cmidrule(l){5-7}
EMG bands & User & Stage & User+Stage & User & Stage & User+Stage \\
\midrule
L & $16.912 \pm 1.595$ & $19.487 \pm 1.730$ & $19.352 \pm 1.724$ & $17.003 \pm 1.635$ & $19.479 \pm 1.717$ & $19.365 \pm 1.737$ \\
M & $16.896 \pm 1.599$ & $19.496 \pm 1.748$ & $19.361 \pm 1.685$ & $16.885 \pm 1.597$ & $19.509 \pm 1.754$ & $19.370 \pm 1.684$ \\
H & $17.040 \pm 1.601$ & $19.514 \pm 1.724$ & $19.536 \pm 1.597$ & $17.054 \pm 1.610$ & $19.515 \pm 1.722$ & $19.543 \pm 1.593$ \\
L+M & $16.699 \pm 1.590$ & $19.438 \pm 1.767$ & $19.267 \pm 1.705$ & $16.695 \pm 1.578$ & $19.433 \pm 1.767$ & $19.272 \pm 1.690$ \\
L+H & $16.692 \pm 1.577$ & $19.408 \pm 1.745$ & $19.281 \pm 1.744$ & $16.722 \pm 1.603$ & $19.399 \pm 1.740$ & $19.295 \pm 1.726$ \\
M+H & $16.828 \pm 1.601$ & $19.434 \pm 1.749$ & $19.323 \pm 1.667$ & $16.831 \pm 1.601$ & $19.434 \pm 1.754$ & $19.324 \pm 1.670$ \\
\addlinespace[2pt]
L+M+H & $16.641 \pm 1.578$ & $19.386 \pm 1.763$ & $19.240 \pm 1.699$ & $16.637 \pm 1.581$ & $19.376 \pm 1.764$ & $19.233 \pm 1.692$ \\
\bottomrule
\end{tabular*}
\end{table*}

%% file: tables/band_information_removal.tex
% Automatically generated from verified paired-user contrasts.
\begin{table*}[t]
\centering
\footnotesize
\setlength{\tabcolsep}{2pt}
\renewcommand{\arraystretch}{1.08}
\caption{Effect of retaining each band, computed as reduced-band minus full-band MAE after independent refitting. Positive $\Delta_b$ indicates a benefit from retaining band $b$. Intervals are marginal paired-user bootstrap 95\% CIs in degrees. The $p_{\mathrm{H}}$ values retain the original Holm adjustment over 18 contrasts per window definition; the correction is not reduced to the nine contrasts displayed here. Confidence intervals are not multiplicity-adjusted.}
\label{tab:band_information_removal}
\begin{tabular*}{\textwidth}{@{\extracolsep{\fill}}ll*{6}{c}@{}}
\toprule
& & \multicolumn{3}{c}{Overlapping windows} & \multicolumn{3}{c}{Disjoint windows} \\
\cmidrule(lr){3-5}\cmidrule(l){6-8}
Condition & Band & $\Delta_b$ & 95\% CI & $p_{\mathrm{H}}$ & $\Delta_b$ & 95\% CI & $p_{\mathrm{H}}$ \\
\midrule
User & L & $0.1875$ & $[0.1048,\,0.2738]$ & $0.0018$ & $0.1939$ & $[0.1028,\,0.2897]$ & $0.0078$ \\
 & M & $0.0516$ & $[0.0277,\,0.0783]$ & $0.0060$ & $0.0852$ & $[0.0435,\,0.1343]$ & $0.0028$ \\
 & H & $0.0579$ & $[0.0266,\,0.0892]$ & $0.0330$ & $0.0585$ & $[0.0040,\,0.1050]$ & $0.2891$ \\
\addlinespace[2pt]
Stage & L & $0.0485$ & $[0.0104,\,0.0851]$ & $0.1490$ & $0.0584$ & $[0.0206,\,0.0946]$ & $0.0264$ \\
 & M & $0.0220$ & $[0.0104,\,0.0337]$ & $0.0039$ & $0.0232$ & $[0.0093,\,0.0371]$ & $0.0180$ \\
 & H & $0.0519$ & $[0.0381,\,0.0664]$ & $0.0018$ & $0.0572$ & $[0.0411,\,0.0748]$ & $0.0018$ \\
\addlinespace[2pt]
User+Stage & L & $0.0833$ & $[-0.0496,\,0.2103]$ & $0.5687$ & $0.0913$ & $[-0.0479,\,0.2211]$ & $0.3966$ \\
 & M & $0.0408$ & $[0.0070,\,0.0749]$ & $0.2674$ & $0.0622$ & $[0.0292,\,0.0945]$ & $0.0264$ \\
 & H & $0.0267$ & $[-0.0127,\,0.0658]$ & $0.5687$ & $0.0385$ & $[-0.0092,\,0.0873]$ & $0.3966$ \\
\bottomrule
\end{tabular*}
\end{table*}

%% file: appendix/appendix_synapmyo_qualitative.tex
% Requires \usepackage{graphicx}.
% Copy synapmyo_qualitative_all_baselines.pdf into figures/.
\subsection{Qualitative Comparison on SynapMyo}
\label{app:synapmyo_qualitative}

\begin{figure*}[p]
    \centering
    \includegraphics[
        width=\textwidth,
        height=0.84\textheight,
        keepaspectratio
    ]{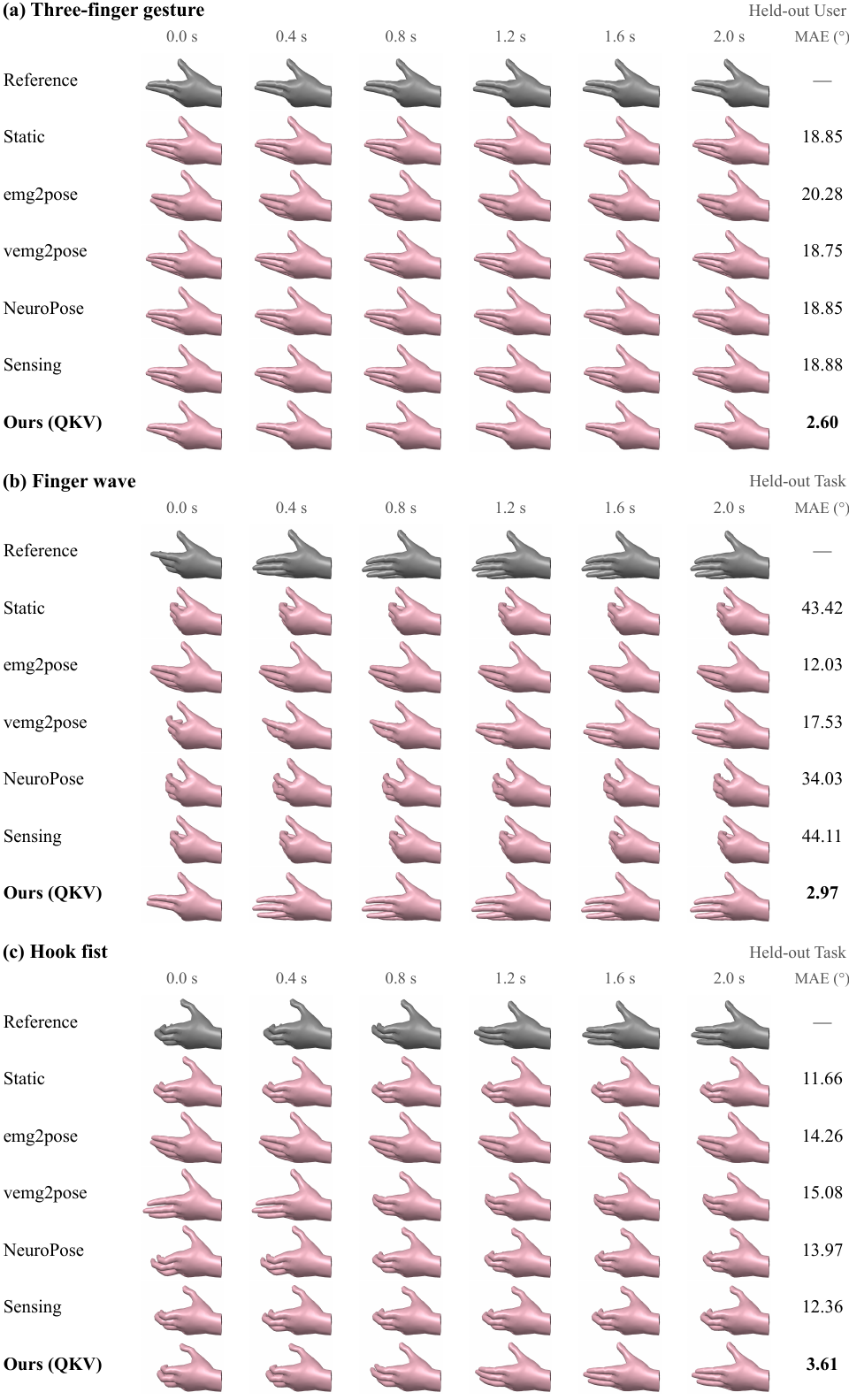}
    \caption{Qualitative Tracking comparison on eight-channel
    SynapMyo with all five baselines from
    Table~\ref{tab:synapmyo_main}:
    (a) a three-finger gesture under held-out User;
    (b) finger wave and (c) hook fist under held-out Task.
    Gray meshes show vision-derived UmeTrack annotations;
    pink meshes show predictions. Static holds the original
    evaluation window's initial pose fixed. Six snapshots span
    each two-second crop; predictions are not reinitialized at
    the crop boundary. All methods share the same hand geometry,
    view, scale, and lighting. The right column reports clip-level
    angular MAE. Ours uses the Table~\ref{tab:synapmyo_main}
    QKV checkpoint. These are selected positive examples.}
    \label{fig:synapmyo_qualitative_all_baselines}
\end{figure*}

Figure~\ref{fig:synapmyo_qualitative_all_baselines} compares
InterBand-EMG (QKV) with all five baselines in
Table~\ref{tab:synapmyo_main} on three selected test clips.
The examples show differences in finger configuration and
transition timing for a three-finger gesture, finger wave,
and hook fist. InterBand-EMG achieves clip-level angular MAEs
of $2.60^\circ$, $2.97^\circ$, and $3.61^\circ$, compared with
$18.75^\circ$, $17.53^\circ$, and $15.08^\circ$ for vemg2pose.
These are selected positive examples in which QKV has lower
MAE, LD, and FD than all five baselines; aggregate performance
is reported in Table~\ref{tab:synapmyo_main}.

